# Learning-based estimation of cattle weight gain and its influencing factors

Muhammad Riaz Hasib Hossain [a, *], Rafiqul Islam [b], Shawn R. McGrath [c], Md Zahidul Islam [d], and David Lamb [e, f]

[a] *School of Computing, Mathematics and Engineering, Charles Sturt University, Wagga Wagga, NSW 2650, Australia*
[b] *School of Computing, Mathematics and Engineering, Charles Sturt University, Albury, NSW 2640, Australia*
[c] *Gulbali Institute for Agriculture, Water and Environment, Charles Sturt University, Wagga Wagga, NSW 2678, Australia*
[d] *School of Computing, Mathematics and Engineering, Charles Sturt University, Bathurst, NSW 2795, Australia*
[e] *Precision Agriculture Research Group, University of New England NSW, Australia 2351*
[f] *Food Agility CRC Ltd, Sydney, NSW, Australia 2000.*

**Abstract**

Many cattle farmers still depend on manual methods to measure the live weight gain of cattle at set intervals, which is time-consuming, labour-intensive, and stressful for both the animals and handlers. A remote and autonomous monitoring system using machine learning (ML) or deep learning (DL) can provide a more efficient and less invasive method and also predictive capabilities for future cattle weight gain (CWG). This system allows continuous monitoring and estimation of individual cattle's live weight gain, growth rates and weight fluctuations considering various factors like environmental conditions, genetic predispositions, feed availability, movement patterns and behaviour. Several researchers have explored the efficiency of estimating CWG using ML and DL algorithms. However, estimating CWG suffers from a lack of consistency in its application. Moreover, ML or DL can provide weight gain estimations based on several features that vary in existing research. Additionally, previous studies have encountered various data-related challenges when estimating CWG. This paper presents a comprehensive investigation in estimating CWG using advanced ML techniques based on research articles (2004-2024). This study investigates the current tools, methods, and features used in CWG estimation, as well as their strengths and weaknesses. The findings highlight the significance of using advanced ML approaches in CWG estimation and its critical influence on factors. Furthermore, this study identifies potential research gaps and provides research direction on CWG prediction, which serves as a reference for future research in this area.

## 1. Introduction

Cattle weight gain (CWG) refers to the increase in an animal's live weight over a specified period and is a critical metric in livestock management. CWG is a crucial indicator for various production outcomes, including milk yield efficiency (Pakrashi et al., 2023; Song, Bokkers et al., 2018), disease prevalence, body condition, reproductive health (Osoro & Wright, 1992; Sousa et al., 2018), and overall well-being (Dikmen et al., 2012). Effective monitoring of CWG is essential for meeting market demands, maximising profitability, and ensuring herd welfare (Bewley, 2010; Eastwood et al., 2016; Fournel et al., 2017; Mc Hugh et al., 2011). Several factors, including genetics, diet, environmental conditions, and management strategies, influence CWG.

In countries like Australia, optimising cattle production efficiency has become a priority. Since CWG directly affects resource allocation, it plays a central role in enhancing productivity and profitability (Halachmi et al., 2019). Therefore, accurate and efficient estimation of weight gain is vital for effective decision-making in areas such as nutrition, health, and performance, ensuring that cattle growth aligns with production targets.

* Corresponding author: Muhammad Riaz Hasib Hossain (email: muhossain@csu.edu.au)

Several methods are available for estimating CWG. A traditional approach is visual assessment, where experienced livestock handlers estimate weight gain based on body condition, muscle development, and size (Wood et al., 2015). Body condition scoring is another method used to estimate CWG by evaluating the animal's fatness and overall health. However, like other manual techniques, this method can be time-consuming, stressful, and costly, often requiring specialised equipment (Brown-Brandl et al., 2010; Weber et al., 2020). Although reasonably accurate, this method requires significant expertise. Another approach involves the use of weight tapes, which measure an animal's girth to estimate its body weight. For greater precision, portable livestock scales or weighbridges are employed, though these methods can be time-consuming (Dang et al., 2022) and stressful (Biase et al., 2022; Cominotte et al., 2020) for the cattle. The stress caused by scales can also lead to a decrease in weight and productivity by approximately 5–10% (Ruchay, Kober, Dorofeev, Kolpakov, Gladkov, et al., 2022). With the global expansion of large-scale cattle farms, it is becoming increasingly difficult to apply these labour-intensive methods to individual animals.

Automated weighing systems have evolved to include advanced technologies, with two primary techniques gaining commercial prominence: walk-over weighing (WoW) and step-on-off (SOO) systems (Duwalage et al., 2023). In the WoW system, the entire animal interacts with the weighing sensor (Parsons et al., 2023; Simanungkalit et al., 2020), whereas the SOO system typically measures only a portion of the animal, such as the front legs ("Optiweigh", 2021). Despite their widespread adoption, they present limitations, including potential environmental damage and infrequent measurement intervals, which may hinder their accuracy and practical application (Bahashwan, 2014; Xiong et al., 2023; Yan et al., 2019). Table 1 outlines various commercial systems that apply WoW and SOO approaches to improve cattle weight monitoring.

**Table 1**
Commercial Cattle Weighing Solutions

| Name | Description | Country | Reference |
|---|---|---|---|
| Cardinal Wrangler Mobile Livestock Scale | A portable weighing solution designed for groups of 15-20 cattle, suitable for use on firm surfaces. | United States of America | ("Weight Wrangler Mobile Livestock Scales", 2024) |
| Gallagher Auto Weigher | This platform attracts animals, identifies them via EID, calculates full body weight, and uploads data to the cloud for access through an app. | Australia | ("Auto Weigher", 2024) |
| Optiweigh | The system includes a portable platform with an RFID reader and a solar-powered satellite for real-time data transmission. It records front-end weight to estimate total weight. | Australia | ("Optiweigh", 2021) |
| Ritchie Beef Monitor | This solution utilises a load-bar platform with cloud-based data access to monitor live weight. | United Kingdom | (Georgina, 2024) |
| Vytelle SENSE | Vytelle SENSE offers a weight-tracking system combined with reproduction and performance monitoring capabilities for comprehensive livestock management. | United States of America | ("Vytelle SENSE", 2025) |

Researchers have used advanced machine learning (ML) techniques to estimate CWG by utilising various features, including cattle image, feed availability, climate, genetic predispositions, animal tracking, and other similar parameters. These techniques provide an alternative to traditional methods by minimising labour costs and reducing stress for both cattle and farmers (Ruchay, Kober, Dorofeev, Kolpakov, Dzhulamanov et al., 2022). In the ML and deep learning (DL) process, each feature appears as a dataset column, functioning as an input for the model (Hossain & Kabir, 2023). Both ML and DL techniques apply mechanisms that identify patterns and gain insights from data, involving training on a dataset, validating on a distinct set, and testing the model's efficacy with a separate test dataset. ML is generally categorised into two major approaches: supervised and unsupervised. When a labelled dataset is used to train an ML model, it falls under supervised ML. Conversely, unsupervised learning involves unlabelled datasets for clustering and analysis via ML algorithms (Ippolito et al., 2021). In contrast, DL methodologies are particularly effective for handling large and complex datasets with high-dimensional features (Ippolito et al., 2021). Consequently, DL models tend to outperform traditional ML models when processing audio, image, speech, text, and video data (Lecun et al., 2015). By utilising existing data, ML and DL algorithms have demonstrated their ability to estimate CWG (Agung et al., 2018; De Vries et al., 1999; Hakem et al., 2021). While the implementation of advanced ML models may seem relatively straightforward, achieving optimal accuracy requires overcoming several challenges. Key challenges include selecting suitable algorithms, fine-tuning parameters, and choosing the most relevant features (Janiesch et al., 2021). Therefore, selecting a proper algorithm is essential for estimating CWG.

In this paper, a systematic literature review (SLR) was performed. Initially, 212 articles were selected from four digital databases (IEEE Xplore, Science Direct, Scopus, and Web of Science) based on the research objectives. Following the sorting process, 53 potential research articles published in recent years that concentrate on estimating the live weight gain of cattle using advanced ML techniques were identified. This study rigorously reviewed all selected articles, highlighting the potential research challenges on CWG estimation and proposing insights into future research directions. The main contributions of this paper are as follows:

- Provide a comprehensive systematic review of academic research on estimating CWG, consolidating insights and methodologies to enhance understanding of the field's trends, variations, and knowledge gaps.

- Identify the potential constraints in previous survey papers on estimating CWG and propose solutions to address these gaps in the literature.

- Explore the complex factors affecting CWG, offering a comprehensive understanding of the variables impacting accuracy while highlighting areas for further research and advancement.

- Assess a diverse range of advanced ML techniques for estimating CWG, providing valuable insights into cattle's weight gain performance across different scenarios and aiding informed decision-making for precise gain estimation.

- Identify and elucidate challenges and limitations, encompassing data quality, data preprocessing, feature selection, environmental variations, and others, laying the groundwork for future research in estimating CWG.

The paper is structured as follows in the subsequent sections: 1) Section 2 delves into the related literature, offering an overview of prior research concerning estimation of CWG; 2) Section 3 outlines the structure of this review; 3) Section 4 encompasses the extracted data from selected publications and presents the corresponding findings addressing the research questions (RQs); 4) Section 5 describes the SLR findings while pointing to the study's future direction; and 5) the concluding section encapsulates the study's results. Fig. 1 presents the overall structure of the study.

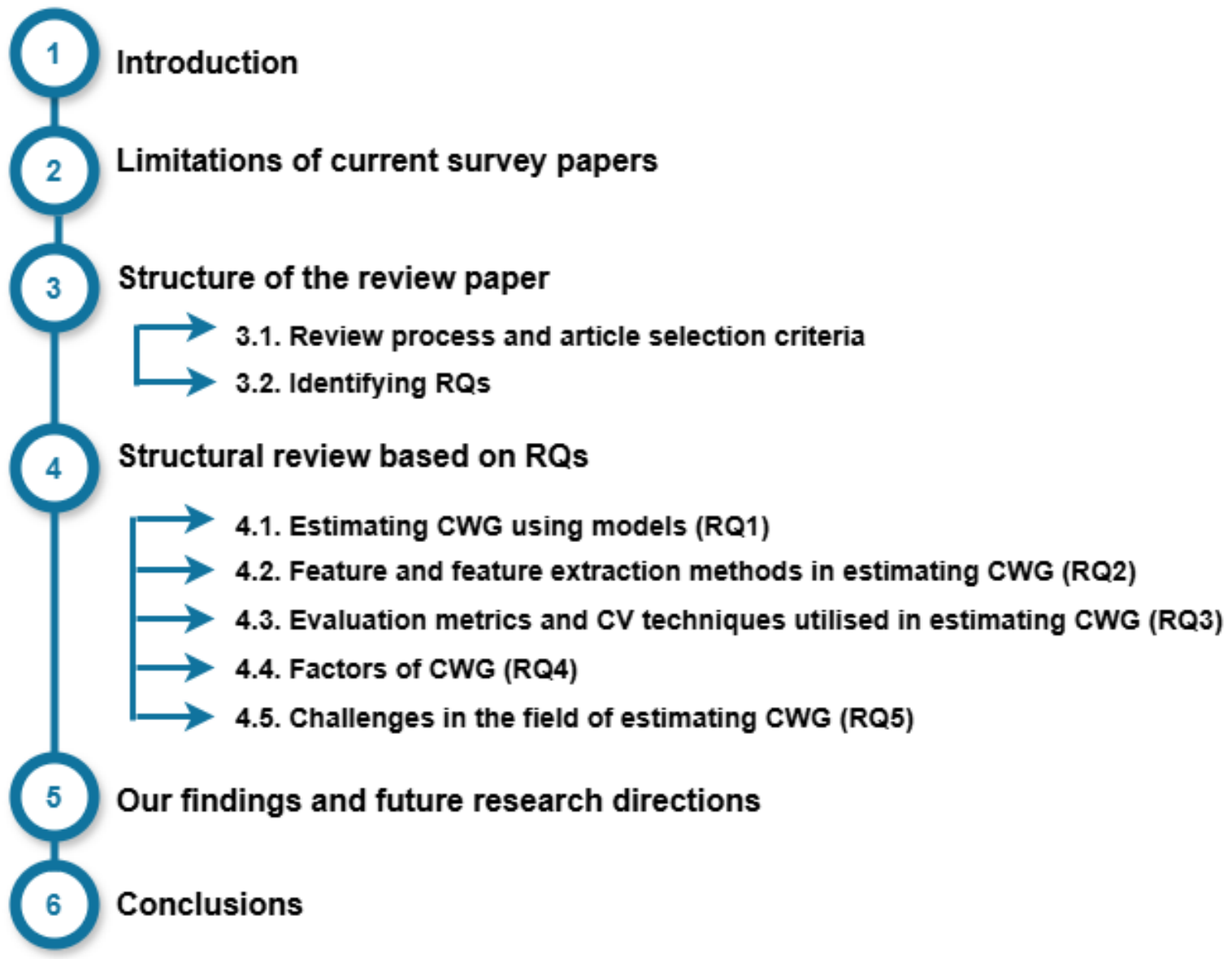

Fig. 1. Overall organization of the paper.

To enhance readability, a list of the abbreviations used in this study is presented in Table 2.

**Table 2**
A list of the abbreviations

| Acronym | Meaning | Acronym | Meaning |
|---|---|---|---|
| AAE | Average Absolute Error | MRE | Mean Relative Error |
| AAPE | Average Absolute Percentage Error | MSE | Mean Square Error |
| ANN | Artificial Neural Networks | PLS | Partial Least Squares |
| BR | Bagging Regression | $r$ | Correlation Coefficient |
| CNN | Convolutional Neural Network | $R^2$ | Coefficient of Determination |
| CV | Cross-Validation | RF | Random Forest |
| CWG | Cattle Weight Gain | RMSE | Root Mean Square Error |
| DL | Deep Learning | RMSEP | Root Mean Square Error of Prediction |
| DT | Decision Tree | RQs | Research Questions |
| K-NN | K-Nearest Neighbor | RD | Regression Discretization |
| LASSO | Least Absolute Shrinkage and Selection Operator | RR | Ridge Regression |
| LightGBM | Light Gradient Boosting Machine | SR | Stepwise Regression |
| LR | Linear Regression | SVM | Support Vector Machine |
| MAE | Mean Absolute Error | SLR | Systematic Literature Review |
| MAPE | Mean Absolute Percentage Error | SOO | Step-On-Off |
| ML | Machine Learning | SSE | Sum Square Error |
| MLP | Multilayer Perceptron | WoW | Walk-Over Weighing |

## 2. Limitations of current survey papers

Researchers have conducted several review studies previously to explore CWG, the factors influencing it, and their estimation processes and techniques. To locate previous SLR papers on estimating CWG, a search strategy was developed for each database (IEEE Xplore, Science Direct, Scopus, and Web of Science). Fifteen literature reviews published from 2004 until October 2024 were retrieved from the databases using search queries. Later, the results were narrowed by eliminating papers unrelated to estimating CWG, excluding redundant review papers across multiple databases, and disregarding non-English publications. Consequently, three relevant SLR papers on estimating CWG were identified. The entire selection process is depicted in Fig. 2.

Three review papers focused on estimating CWG were identified. Although the search strategy included publications from 2004 to 2024 to capture recent survey papers on CWG estimation, the earliest relevant review paper appeared in 2021. Notably, the authors Wang et al. (2021) and Dohmen et al. (2022) conducted exhaustive reviews of research articles that focused on the utilisation of computer vision and advanced ML techniques for the estimation of body weight gain in diverse livestock species. The authors Wang et al. (2021) conducted a comprehensive survey encompassing a wide range of animals, including camels, cattle, goats, pigs, sheep, and yaks, while the authors Dohmen et al. (2022) concentrated their review on cattle and pigs. On the contrary, the authors Qiao et al. (2021) focused their attention on articles specifically addressing cattle. However, their review encompassed a diverse array of subjects, including cattle identification, assessment of body condition scores, and estimation of body weight. Despite addressing specific aspects of estimating CWG, these three review papers demonstrated limitations of comprehensive coverage and often neglected significant research gaps identified within the field. Table 3 summarises the limitations of the existing review papers.

This examination has exposed various deficiencies and constraints in the current body of literature. Previous reviews have fallen short in delivering a comprehensive SLR that covers all relevant scientific journal articles related to estimating CWG. This paper addresses these gaps by conducting a comprehensive SLR of the most up-to-date scientific journal articles that leverage advanced ML techniques to assess CWG and the factors affecting it.

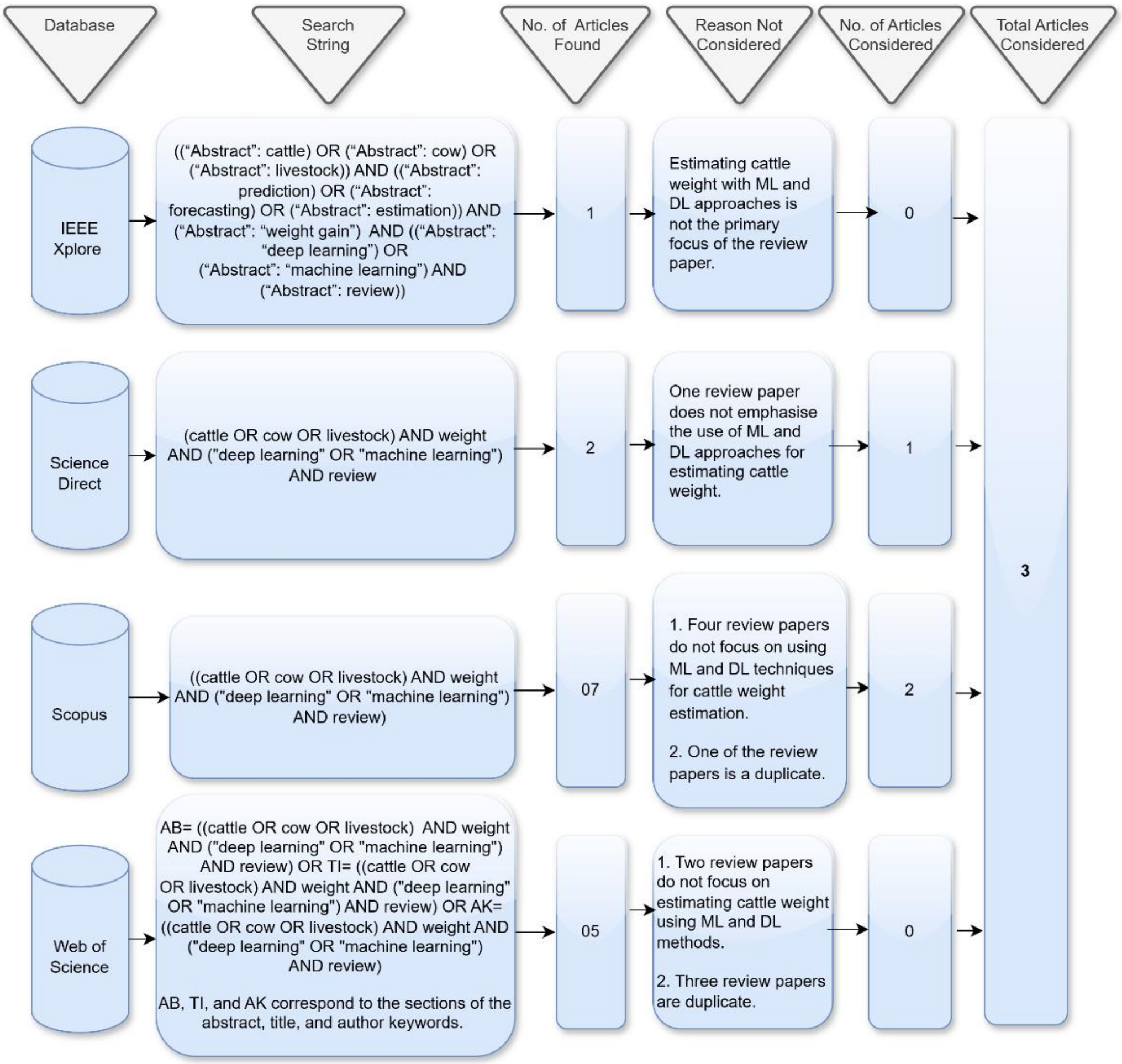

Fig. 2. A flowchart to retrieve relevant literature reviews (2004-2024).

**Table 3**
A list of review papers published between 2004 and 2024

| No | Reference | Focus | Key Observations | Publication Year |
|---|---|---|---|---|
| 1 | Dohmen et al. (2022) | Reviewed 26 selected papers with a concentration on – <br>• Computer vision and ML techniques <br>• Estimating the weight of livestock, especially cattle and pigs | • The focus is on estimating the body weight of cattle and pigs using 2D or 3D sensors. <br>• Analysing both animals together, the outcomes of cattle weight estimation were confusing. | 2022 |
| 2 | Qiao et al. (2021) | Studied 100 selected papers with a highlight on – <br>• Cattle identification, body condition scoring, and body weight estimation | • Concentrate only on studies that used 2D and 3D sensors to estimate cattle body weight. | 2021 |
| 3 | Wang et al. (2021) | Analysed several papers with a focus on – <br>• Computer vision and ML techniques <br>• Estimating the weight of livestock, mainly camels, cattle, goats, pigs, sheep, and yaks | • The target is to estimate the body weight of camels, cattle, goats, pigs, sheep, and yaks using computer vision. <br>• Limited exploration of CWG estimation. | 2021 |

## 3. Structure of the review paper

### *3.1. Review process and article selection criteria*

This paper employs SLR to systematically gather, evaluate, and interpret all relevant published papers on estimating CWG. The goal is to pinpoint potential research gaps, formulate RQs, and provide comprehensive information to researchers. A structured approach was adopted to perform an SLR on estimating CWG. To ensure a structured approach, the SLR process commenced by scrutinising the existing literature to identify knowledge gaps, which formed the foundation for crafting the RQs. Subsequently, customised search strings were used in the study for each digital database, encompassing prominent platforms such as IEEE Xplore, Science Direct, Scopus, and Web of Science. These search strings incorporated carefully selected keywords to facilitate the retrieval of pertinent research articles. Selecting appropriate search terms, as well as implementing a coherent search strategy, are crucial for extracting relevant papers from digital databases. This strategic approach was designed to explore the subject matter within the cattle industry comprehensively. After establishing the RQs, a tailored search strategy was developed for each database, with the specific details outlined in Table 4. This critical phase was essential for ensuring a comprehensive survey of the pertinent literature within the cattle industry. The execution of the formulated search queries resulted in the successful retrieval of 212 journal articles, playing a crucial role in addressing the defined research questions and making a significant contribution to a comprehensive understanding of the subject matter.

**Table 4**
Search strings for four databases to retrieve scientific journal articles

| SN | Database Name | Search String | Number of Articles Found |
|---|---|---|---|
| 01 | IEEE Xplore | ((“Abstract”: cattle) OR (“Abstract”: cow) OR (“Abstract”: livestock)) AND ((“Abstract”: prediction) OR (“Abstract”: forecasting) OR (“Abstract”: estimation)) AND (“Abstract”: “weight gain”) AND ((“Abstract”: “deep learning”) OR (“Abstract”: “machine learning”)) | 02 |
| 02 | Science Direct | (cattle OR cow OR livestock) AND “weight gain” AND (estimation OR prediction OR forecasting) AND (“deep learning” OR “machine learning”)<br><br>The search string exclusively scans the database’s title, abstract, and keyword sections. | 77 |
| 03 | Scopus | TITLE-ABS-KEY ((cattle OR cow OR livestock) AND “weight gain” AND (estimation OR prediction OR forecasting) AND (“deep learning” OR “machine learning”))<br><br>The search string exclusively scans the database’s title, abstract, and keyword sections. | 85 |
| 04 | Web of Science | AB= ((cattle OR cow OR livestock) AND “weight gain” AND (“deep learning” OR “machine learning”)) OR TI= ((cattle OR cow OR livestock) AND “weight gain” AND (“deep learning” OR “machine learning”)) OR AK= ((cattle OR cow OR livestock) AND weight AND (“deep learning” OR “machine learning”))<br><br>AB, TI, and AK correspond to the abstract, title, and author keywords section. | 48 |

The eligibilities for selection are employed to determine which research articles have the potential to address the RQs. In this investigation, the inclusion and exclusion criteria derived from the RQs were implemented. A spreadsheet assessed the inclusion and exclusion criteria to monitor the outcomes of the search strings from the databases. In the SLR, research articles that satisfied the inclusion criteria and did not meet the exclusion criteria were included. The exclusion criteria involved excluding publications unrelated to ML or DL in the context of estimating CWG and those not written in English. In contrast, publications were considered in the inclusion criteria if those utilised ML or DL approaches for estimating CWG. After removing redundant records, the eligibility criteria were employed for the remaining papers, resulting in the evaluation of 39 full-text articles. As part of the selection process, a quality assessment of the chosen articles was conducted following the guidelines by Kitchenham et al. (2009). Furthermore, the technique of backward and forward snowballing (Wohlin, 2014) was employed, leading to the identification of twelve more relevant articles. As a result, 53 articles were uncovered for the SLR. The entire selection process is depicted in Fig. 3.

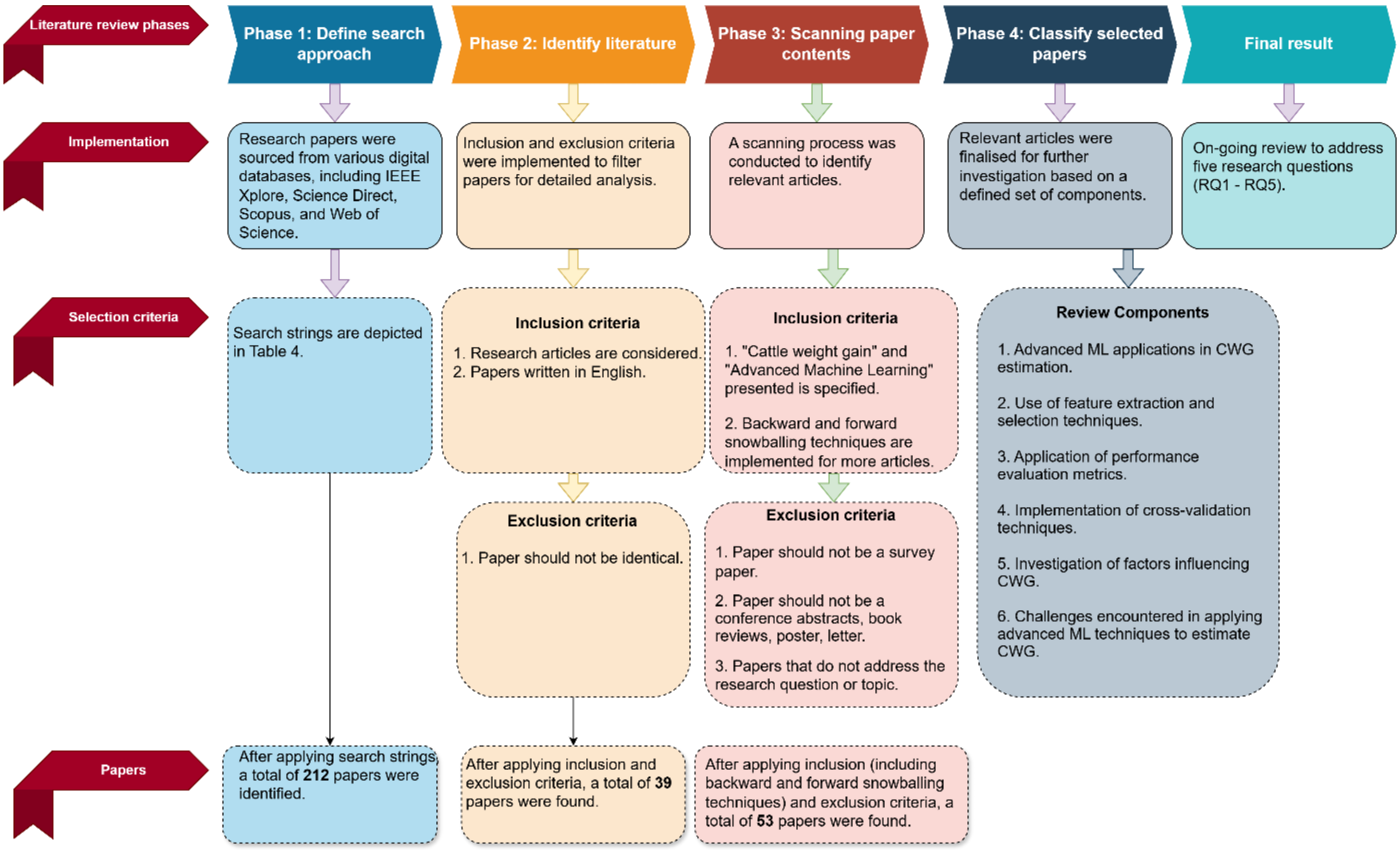

Fig. 3. Papers selection process for the review (1992-2024).

### 3.2. Identifying RQs

Defining appropriate RQs is essential as they guide the acquisition of knowledge. Hence, the initial step in the SLR involves formulating RQs that address the current state of research on estimating CWG. Consequently, the SLR has identified five RQs and their respective motivations, as shown in Table 5.

Fig. 4 offers a visual roadmap illustrating the interconnections among 53 research articles centred on estimating CWG. Such visual representations are vital tools for discerning and analysing emerging trends within the field, pinpointing gaps in the existing literature, and tracing the evolution of ideas and concepts unique to estimating CWG. This visual mapping not only enhances comprehension of the subject matter but also facilitates the identification of key research themes and areas that require further exploration. It is a valuable resource for researchers and scholars seeking to navigate the landscape of estimating CWG literature and make informed contributions to this dynamic field.

**Table 5**
List of the identified RQs

| SN | RQ | Main Motivation |
|---|---|---|
| RQ1 | What is the trend of ML and DL models in the literature for estimating CWG? | 1. To provide an up-to-date analysis of the latest advancements in ML and DL techniques for estimating CWG.<br>2. To evaluate the effectiveness and reliability of ML and DL models in the context of estimating CWG.<br>3. To identify areas within the existing literature where further research is needed.<br>4. To offer insights and guidance to researchers and practitioners interested in using ML and DL for estimating CWG. |
| RQ2 | Which features and feature extraction methods have researchers employed to utilise ML and DL for estimating CWG? | 1. To evaluate the range of features and feature extraction approaches within ML and DL models for estimating CWG.<br>2. To compare different feature sets and extraction methods to determine their effectiveness.<br>3. To identify areas where there may be a lack of research or exploration regarding feature engineering for estimating CWG.<br>4. To contribute to refining models by understanding which features and methods are most effective. |
| RQ3 | What are the performance evaluation metrics and CV techniques utilised in the literature for estimating CWG? | 1. To assess the performance evaluation metrics and CV techniques employed in the context of estimating CWG.<br>2. To provide an up-to-date overview of the evaluation metrics and CV techniques currently used in this field.<br>3. To enable comparisons of different performance evaluation metrics to determine suitability and effectiveness. |
| RQ4 | What factors have researchers discovered for CWG? | 1. To understand the factors that influence CWG can contribute to improved animal health and greater efficiency in livestock production. |
| RQ5 | What difficulties or challenges arise when using ML and DL models to estimate CWG? | 1. To evaluate the obstacles and issues when utilising ML and DL models for estimating CWG.<br>2. To provide a consolidated understanding of the common obstacles researchers and practitioners face in this field.<br>3. To pinpoint areas where further research is needed to overcome specific challenges. |

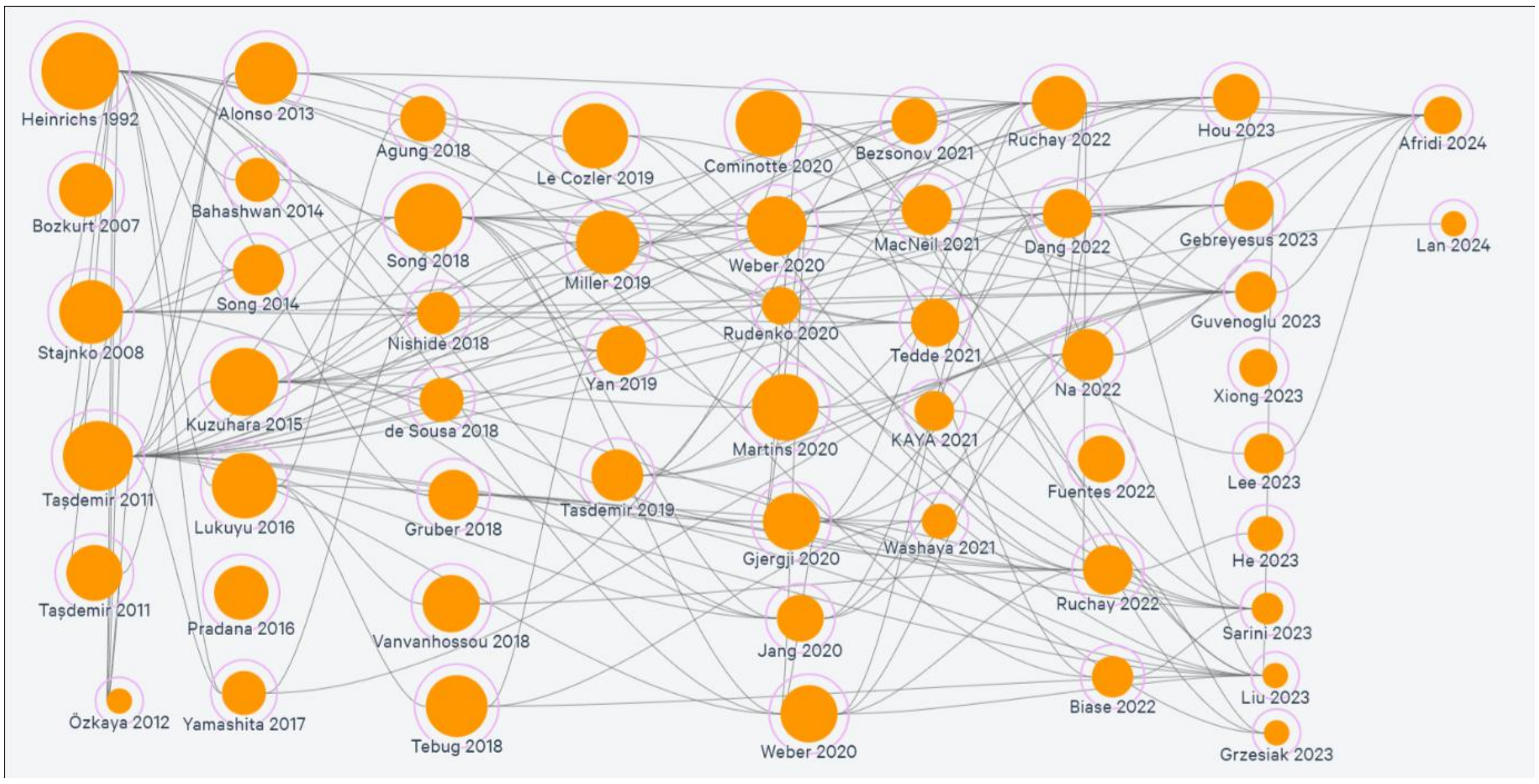

Fig. 4. Literature mapping of the selected papers.

## 4. Structural review based on RQs

This section presents the extracted data from the selected papers, providing answers to research questions. It highlights fifty-one research articles, forming a compilation of recent papers focused on estimating CWG.

### *4.1. Estimating CWG using models (RQ1)*

Advanced ML techniques have demonstrated impressive performance in estimating CWG in livestock farming. Table 6 provides a comprehensive summary of 53 selected papers, encompassing details such as the top-performing model (along with the source code/ tools used), its accuracy, dataset size (including types), and the application of cross-validation (CV) in recent publications on estimating CWG.

**Table 6**
An assessment of ML and DL models

| SN | Paper Reference | Database | | Best / Proposed Model | Source Code/tools used in the AI model | Best Result |
|---|---|---|---|---|---|---|
| | | Size | CV | | | |
| 01. | Afridi et al. (2024) | 1289(*I*) | Holdout (80%, -, 20%) | CNN | - | $R^2$ = 0.97 |
| 02. | Agung et al. (2018) | 63 (*N*) | - | LR | - | $R^2$ = 0.84 |
| 03. | Alonso et al. (2013) | 390 (*N*) | 10-fold | SVM | - | MAPE = 4.27% |
| 04. | Bahashwan et al. (2014) | 108 (*N*) | - | LR | - | $R^2$ = 0.915 |
| 05. | Bezsonov et al. (2021) | 250 (*I*) | - | MLP | - | Accuracy = 92% |
| 06. | Biase et al. (2022) | 71 (*N*) | Leave-One-Out | ANN | - | MAPE = 1.3769% |
| 07. | Bozkurt et al. (2007) | 140 (*I*) | - | LR | - | $R^2$ = 63.6% |
| 08. | Cominotte et al. (2020) | 234 (*I*) | Leave-One-Out | ANN | Software R | $R^2$ = 0.91 |
| 09. | Dang et al. (2022) | 33,546 (*N*) | Holdout (70%, 15%, 15%) | LightGBM | Python | RMSE = 24.754 |
| 10. | de Moraes Weber et al. (2020) | 68 (*I*) | Holdout (66%, -, 34%) | LR and SVM | - | $R^2$ = 0.70 |
| 11. | Fuentes et al.(2022) | 282 (*I*) | Holdout (70, -, 15) | ANN | Matlab using Jupyter Notebook | $R$ = 0.96 |
| 12. | Gebreyesus et al. (2023) | 83,011 (*I*) | Holdout (80%, -, 20%) | RF and DT | Python | $R$ = 0.94 |
| 13. | Gjergji et al. (2020) | 20 (*I*) | Holdout (60%, 20%, 20%) | CNN | PyTorch | MAE = 23.19 kg |
| 14. | Gruber et al. (2018) | 44441 (*N*) | - | LR | SAS 9.4 software | RMSE = 50 kg |
| 15. | Grzesiak et al.(2023) | 840 (*N*) | 10-fold | RF | - | Accuracy = 0.83 |
| 16. | Guvenoglu et al. (2023) | 340 (*I*) | 10-fold | ANN | Python | Error = ± 20 kg |
| 17. | He et al. (2023) | 626 (*N*) | Holdout (501, -, 125) | SVM | - | RMSE = 4.83 kg |
| 18. | Heinrichs et al. (1992) | 2625 (*N*) | - | LR | - | $R^2$ = 0.95 |
| 19. | Hou et al. (2023) | 2106(*I*) | Holdout (1900, 100, 106) | PointNet++ of DL | Python 3.6 using PyTorch and TensorFlow frameworks | RMSE =10.2 kg |
| 20. | Jang et al. (2020) | 209 (*I*) | 5-fold, 10-fold, and so on | LR | Matlab | $R^2$ = 0.9573 |
| 21. | Kaya & Bardakcioglu (2021) | 204 (*I*) | - | SR | Matlab | $R^2$ = 0.75 |
| 22. | Kuzuhara et al. (2015) | 91 (*I*) | - | LR | Software R 3.0.3 | $R^2$ = 0.80 |
| 23. | Lan et al. (2024) | 1800 (*I*) | Holdout (1500, -, 300) | CNN | - | RMSE =15.57 kg |
| 24. | Le Cozler et al. (2019) | 177 (*I*) | 10-fold | LR | Software R | $R^2$ = 0.93 |
| 25. | Lee et al. (2023) | 9359 (*I*) | Holdout (4251, -, 4765) | CNN | TensorFlow 2.5.0, PyTorch 1.10.0 | MAPE = 5.52% |
| 26. | Liu et al. (2023) | 505 (*I*) | - | RF | Python | MAPE = 4.1% |
| 27. | Lukuyu et al. (2016) | 452 (*N*) | - | SR | - | $R^2$ = 0.705 |
| 28. | MacNeil et al. (2021) | 8972 (*N*) | - | LR | - | $R$ = 0.949 |
| 29. | Martins et al.(2020) | 55 (*I*) | - | LR | Matlab | $R^2$ = 0.96 |
| 30. | Miller et al. (2019) | 17127 (*I*) | 10-fold | ANN | Software R | $R^2$ = 0.7 |
| 31. | Na et al. (2022) | 353 (*I*) | - | RF | - | $R^2$ = 0.969 |
| 32. | Nishide et al. (2018) | 184 (*I*) | - | LR | - | $R$ = 0.9107 |
| 33. | Ozkaya (2013) | 41 (*I*) | - | LR | - | $R^2$ = 0.86 |
| 34. | Pradana et al.(2017) | - | - | LR | - | Accuracy = 73.21% |
| 35. | Ruchay, Kober, Dorofeev, Kolpakov, Gladkov et al. (2022) | 59700 (*I*) | Holdout (50%, 20%, 30%) | CNN | - | Accuracy = 91.6% |
| 36. | Ruchay, Kober, Dorofeev, Kolpakov, Dzhulamanov et al. (2022) | 1523 (*N*) | Holdout (50%, 20%, 30%) | LR | SciKit-Learn library | $R^2$ = 0.995 |
| 37. | Rudenko et al. (2020) | 250 (*I*) | - | CNN | - | Accuracy = 92% |
| 38. | Sarini et al. (2023) | 439(*N*) | Holdout (60%, 20%, 10%) | LR for female cattle | - | RMSE = 9.6172 |
| 39. | Song et al. (2014) | 68 (*I*) | Holdout (49, -, 19) | LR | Matlab | $R^2$ = 0.91 |
| 40. | Song et al. (2018) | 30 (*I*) | Leave-One-Out | LR | Matlab | RMSE = 41.2kg |
| 41. | Sousa et al. (2018) | 304 (*I*) | Holdout (80%, 10%, 10%) | ANN | Matlab | $R^2$ = 0.85 |
| 42. | Stajnko et al. (2008) | 12 (*I*) | - | LR | SPSS software 11.0 | $R^2$ = 0.74 |
| 43. | Tasdemir et al. (2011a) | - | - | FM | - | R = 0.99 |
| 44. | Tasdemir et al. (2011b) | 115 (*I*) | - | LR | Matlab | $R^2$ = 0.958 |
| 45. | Tasdemir & Ozkan (2019) | 575 (*I*) | Holdout (75%, -, 25%) | ANN | Matlab | $R$ = 0.995 |
| 46. | Tebug et al. (2018) | 459 (*N*) | - | SR | - | $R^2$ = 0.85 |
| 47. | Tedde et al. (2021) | 1849 (*N*) | 10-fold | PLS | Software R | $R^2$ = 0.61 |
| 48. | Vanvanhossou et al. (2018) | 289 (*N*) | - | LR | SAS 9.2 software | $R^2$ = 0.97 |
| 49. | Wesly et al. (2021) | 1085 (*N*) | - | LR | - | $R^2$ = 0.79 |
| 50. | Weber et al. (2020) | 110 (*I*) | 10-fold | BR | - | $R$ = 0.75 |
| 51. | Xiong et al. (2023) | 53 (*I*) | Holdout (80%, -, 20%,) and 5-fold | BR | Matlab | $R^2$ = 0.83 |
| 52. | Yamashita et al. (2017) | 240 (*I*) | - | LR | - | $R^2$ = 0.8679 |
| 53. | Yan et al. (2019) | 146 (*I*) | Leave-One-Out | LR | SPSS software 19.0 | $R^2$ = 0.972 |

Here, *I* = images, *N* = numerical data, ANN = Artificial Neural Networks, BR = Bagging Regression, CNN = Convolutional Neural Networks, DL = Deep Learning, DT = Decision Tree, FM = Fuzzy Model, LightGBM = Light Gradient Boosting Machine, LR = Linear Regression, MLP = Multilayer Perceptron, PLS = Partial Least Squares, RF = Random Forest, SR = Stepwise Regression, and SVM = Support Vector Machine.

In the extensive examination of 53 academic papers, it has come to attention that researchers have employed a variety of advanced ML technologies to estimate CWG. These technologies encompass Artificial Neural Networks (ANN), Bagging Regression (BR), Bayesian Ridge, Convolutional Neural Network (CNN), Decision Tree (DT), Fuzzy, K-nearest Neighbor (K-NN), Least Absolute Shrinkage and Selection Operator (LASSO), Light Gradient Boosting Machine (LightGBM), Linear Regression (LR), Multilayer Perceptron (MLP), Partial Least Squares (PLS), PointNet++, Random Forest (RF), Regression Discretization (RD), Ridge Regression (RR), Stepwise Regression (SR), and Support Vector Machine (SVM). The findings, as presented in Table 7, offer a comprehensive overview of these ML and DL models, outlining their strengths and weaknesses, usage frequency and the year of their introduction in the context of estimating CWG. As per the table, LR is the most used ML model, employed 31 times for estimating CWG research and consistently has been applied since 1992. Following are ANN and SVM, employed nine times each. Notably, SVM was introduced in 2013, while ANN was incorporated into the analyses starting in 2018.

**Table 7**
ML and DL models adoption for estimating CWG

| SN | ML and DL | References (most cited papers) | Number of papers | Starting year | Strengths and weaknesses |
|---|---|---|---|---|---|
| 01 | LR | Bahashwan (2014), Bozkurt et al. (2007), Cominotte et al.(2020), de Moraes Weber et al. (2020), Gebreyesus et al. (2023), Gruber et al. (2018), Heinrichs et al. (1992), Jang et al. (2020), Kuzuhara et al. (2015), Le Cozler et al. (2019), Na et al. (2022), Ozkaya (2013), Pradana et al. (2017), Ruchay, Kober, Dorofeev, Kolpakov, Dzhulamanov, et al. (2022), Sarini et al. (2023), Song et al. (2014), Song et al. (2018), Stajnko et al. (2008), Tasdemir et al. (2011b), and more | 31 | 1992 | Strengths of the LR:<br>1. Efficient for making predictions with clear linear correlation.<br>2. Simple to interpret.<br><br>Weaknesses of the LR:<br>1. Performance may not be well with complex, non-linear relationships.<br>2. Sensitivity to outliers can impact accuracy, especially in specific scenarios. |
| 02 | Fuzzy | Tasdemir et al. (2011a) | 01 | 2011 | Strengths of the Fuzzy:<br>1. Manage uncertainty and imprecision in data effectively.<br><br>Weaknesses of the Fuzzy:<br>1. Sensitive to input variations.<br>2. Require precise parameter adjustments for optimal performance. |
| 03 | SVM | Alonso et al. (2013), de Moraes Weber et al. (2020), Na et al. (2022), and more | 09 | 2013 | Strengths of the SVM:<br>1. Effectively manage complex data.<br>2. Robust performance with limited labelled training data.<br><br>Weaknesses of the SVM:<br>1. Struggle with extensive datasets.<br>2. Computationally intensive, reducing efficiency. |
| 04 | SR | de Moraes Weber et al. (2020), Kaya & Bardakcioglu (2021), Lukuyu et al. (2016), and Tebug et al. (2018) | 04 | 2016 | Strengths of the SR:<br>1. Automatic variable selection.<br>2. Simplified modelling process.<br><br>Weaknesses of the SR:<br>1. Risk of overfitting with too many variables.<br>2. Require cautious interpretation of results. |
| 05 | ANN | Cominotte et al. (2020), Miller et al. (2019), Tasdemir & Ozkan (2019), and more | 09 | 2018 | Strengths of the ANN:<br>1. Learn complex structures and correlations in data.<br>2. Effective in prediction tasks.<br><br>Weaknesses of the ANN:<br>1. Challenge to interpret.<br>2. Require a substantial training dataset. |
| 06 | RD | Weber et al. (2020) | 01 | 2020 | Strengths of the RD:<br>1. Simplify complex relationships.<br>2. Categorise continuous variables into discrete categories.<br><br>Weaknesses of the RD:<br>1. Potential loss of information due to discretization.<br>2. Increased complexity in interpreting the model. |
| 07 | BR | Weber et al. (2020) | 01 | 2020 | Strengths of the BR:<br>1. Reduce model variabilities.<br>2. Improve generalization. |
| | | | | | 3. Effective in handling high variance and overfitting.<br><br>Weaknesses of the BR:<br>1. Limited improvement if the underlying model is already stable.<br>2. The choice of the base predictor influences performance. |
| 08 | PLS | Cominotte et al. (2020), and Tedde et al. (2021) | 02 | 2020 | Strengths of the PLS:<br>1. Handle multicollinearity effectively.<br>2. Suitable for high-dimensional datasets.<br>3. Extract relevant information and reduce dimensionality.<br><br>Weaknesses of the PLS:<br>1. Struggle with non-linear relationships.<br>2. Performance may be compromised if predictors and outcomes are not linearly related. |
| 09 | CNN | Bezsonov et al. (2021), Gjergji et al. (2020), Lee et al. (2023), Na et al. (2022), Ruchay, Kober, Dorofeev, Kolpakov, Gladkov, et al. (2022), Afridi et al. (2024),(Lan et al., 2024), and Rudenko et al. (2020) | 08 | 2020 | Strengths of the CNN:<br>1. Autonomously acquire hierarchical features.<br>2. Effective in images and pattern recognition tasks.<br><br>Weaknesses of the CNN:<br>1. Prone to overfitting.<br>2. Require careful tuning to mitigate overfitting. |
| 10 | RF | de Moraes Weber et al. (2020), Gebreyesus et al. (2023), Na et al. (2022), Weber et al. (2020), and more | 07 | 2020 | Strengths of the RF:<br>1. Handle complex relationships well.<br>2. Provide feature importance.<br>3. Resist overfitting effectively.<br><br>Weaknesses of the RF:<br>1. The model can be computationally intensive.<br>2. Ensemble nature makes interpretation more challenging. |
| 11 | LASSO | Cominotte et al. (2020), Gebreyesus et al. (2023), and Na et al. (2022) | 03 | 2020 | Strengths of the LASSO:<br>1. Effective in handling high-dimensional data.<br>2. Encourage sparsity in the model.<br>3. Select relevant features efficiently.<br><br>Weaknesses of the LASSO:<br>1. Struggle with multicollinearity.<br>2. Careful tuning of the regularization parameter is required for optimal performance. |
| 12 | MLP | Bezsonov et al. (2021), Biase et al. (2022), and Dang et al. (2022) | 03 | 2021 | Strengths of the MLP:<br>1. Strong in handling complex patterns.<br>2. Effective with large datasets.<br>3. Learn intricate representations well.<br><br>Weaknesses of the MLP:<br>1. Require tuning of settings.<br>2. Computationally intensive training.<br>3. The model is not ideal for situations with limited computing capabilities. |
| 13 | LightGBM | Dang et al. (2022) | 01 | 2022 | Strengths of the LightGBM:<br>1. Handle large datasets effectively. |
| | | | | | 2. Efficiently processes categorical features.<br>3. Well-suited for predictive tasks in diverse domains.<br><br>Weaknesses of the LightGBM:<br>1. Sensitivity to hyperparameter tuning.<br>2. Careful optimization is required for optimal performance in different scenarios. |
| 14 | DT | Gebreyesus et al. (2023), Na et al. (2022), and more | 05 | 2022 | Strengths of the DT:<br>1. Simplicity and interpretability.<br>2. Handle both quantitative and qualitative data effectively.<br><br>Weaknesses of the DT:<br>1. Prone to overfitting.<br>2. Generalization performance can be impacted, especially with noisy data or deep trees. |
| 15 | Bayesian Ridge | Na et al. (2022) | 01 | 2022 | Strengths of the Bayesian Ridge:<br>1. Manage multicollinearity effectively.<br>2. Provide uncertainty estimates for predictions.<br><br>Weaknesses of the Bayesian Ridge:<br>1. Heightened computational complexity.<br>2. Susceptibility to the selection of prior distributions, impacting performance. |
| 16 | K-NN | Dang et al. (2022), Sarini et al. (2023), and Xiong et al. (2023) | 03 | 2022 | Strengths of the K-NN:<br>1. Straightforward for prediction with clear data patterns.<br>2. Adapt to various types of relationships.<br><br>Weaknesses of the K-NN:<br>1. Sensitivity to outliers.<br>2. Dependence on the choice of the number of neighbours (k).<br>3. The model is expensive, particularly with large datasets. |
| 17 | RR | Gebreyesus et al. (2023), and Na et al. (2022) | 02 | 2022 | Strengths of the RR:<br>1. Effective in handling multicollinearity.<br>2. Prevent overfitting with regularization.<br><br>Weaknesses of the RR:<br>1. The model may not perform well when a sparse model is desired.<br>2. Careful consideration of the regularization parameter is required for optimal results. |
| 18 | PointNet ++ | Hou et al. (2023) | 01 | 2023 | Strengths of the PointNet++:<br>1. Effective in tasks like object recognition and segmentation with point cloud data.<br><br>Weaknesses of the PointNet++:<br>1. Sensitivity to input variations.<br>2. Difficulties in managing irregularly sampled point clouds.<br>3. Performance may be impacted in specific scenarios. |

*4.2. Feature and feature extraction methods in estimating CWG (RQ2)*

In estimation, a feature represents a quantifiable characteristic of an object. As part of the analysis, ML algorithms can generate an estimation of CWG by considering multiple features, such as cattle Body Length, Hip Height, and others. ML has a mechanism to determine patterns and correlations and discover knowledge from datasets. In examining the 53 articles, it is found that researchers commonly employed two categories of morphological features for estimating CWG: image-based morphological features and quantitative morphological features. Numerous researchers incorporated external features alongside morphological features (quantitative or image-based) while estimating CWG. This review provides a comprehensive summary of the features utilised in research articles concerning the estimation of CWG. Within this context, most research considers cattle Body Weight as the dependent variable, while various attributes like Body Length, Hip Height, Heart Girth, and other parameters serve as independent variables (Agung et al., 2018; Sarini et al., 2023). Notably, in DL, it is not necessary to identify these features separately, as the model autonomously learns them from the data (Cang et al., 2019). Fig. 5 shows the ML and DL workflows for estimating CWG.

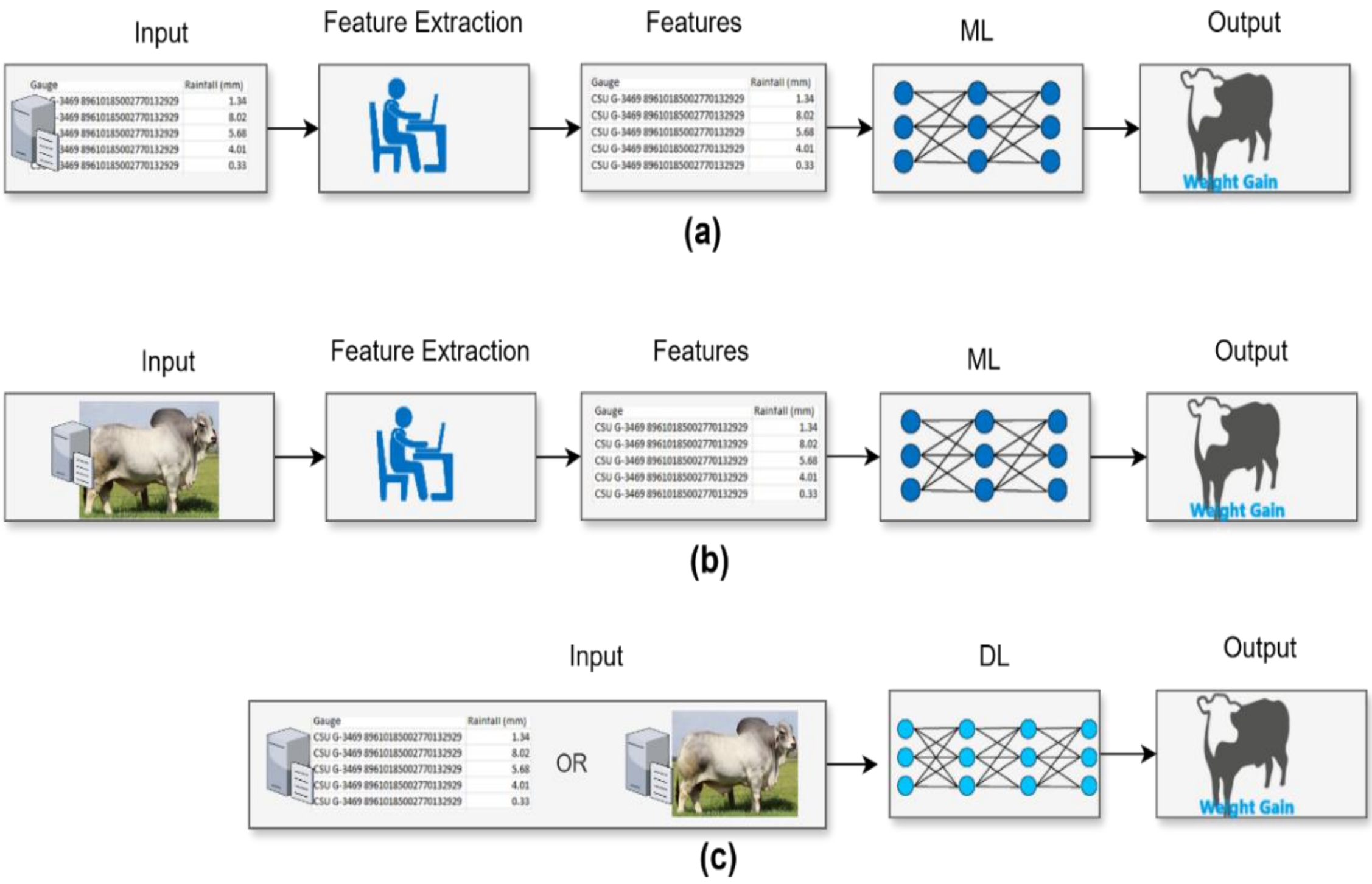

Fig. 5. (a) Quantitative ML workflow. (b) Image-based ML workflow. (c) DL workflow.

Fig. 6 illustrates two distinct technological approaches for cattle weight estimation: image-based setup and numerical data-based setup. Image-based methods employ advanced camera systems to capture multi-angle visuals of cattle, which are then analysed using AI algorithms like deep learning. For instance, RGB-D imagery paired with regression models can achieve over 91% accuracy in weight prediction (Ruchay, Kober, Dorofeev, Kolpakov, Gladkov, et al., 2022). Similarly, ANN models trained on body dimension metrics derived from images report near-perfect correlations (r = 0.99) for weight estimation (Tasdemir & Ozkan, 2019). Recent advancements include the use of supervised learning techniques applied to 3D digital images, where tree-based algorithms such as CatBoost and RF demonstrated high predictive accuracy (r = 0.94) when combining multi-breed datasets (Gebreyesus et al., 2023). Numerical data-based setups, such as automated SOO systems, use embedded sensors to record weight as cattle traverse platforms, offering a hands-free, consistent method for monitoring short-term weight fluctuations (Duwalage et al., 2023).

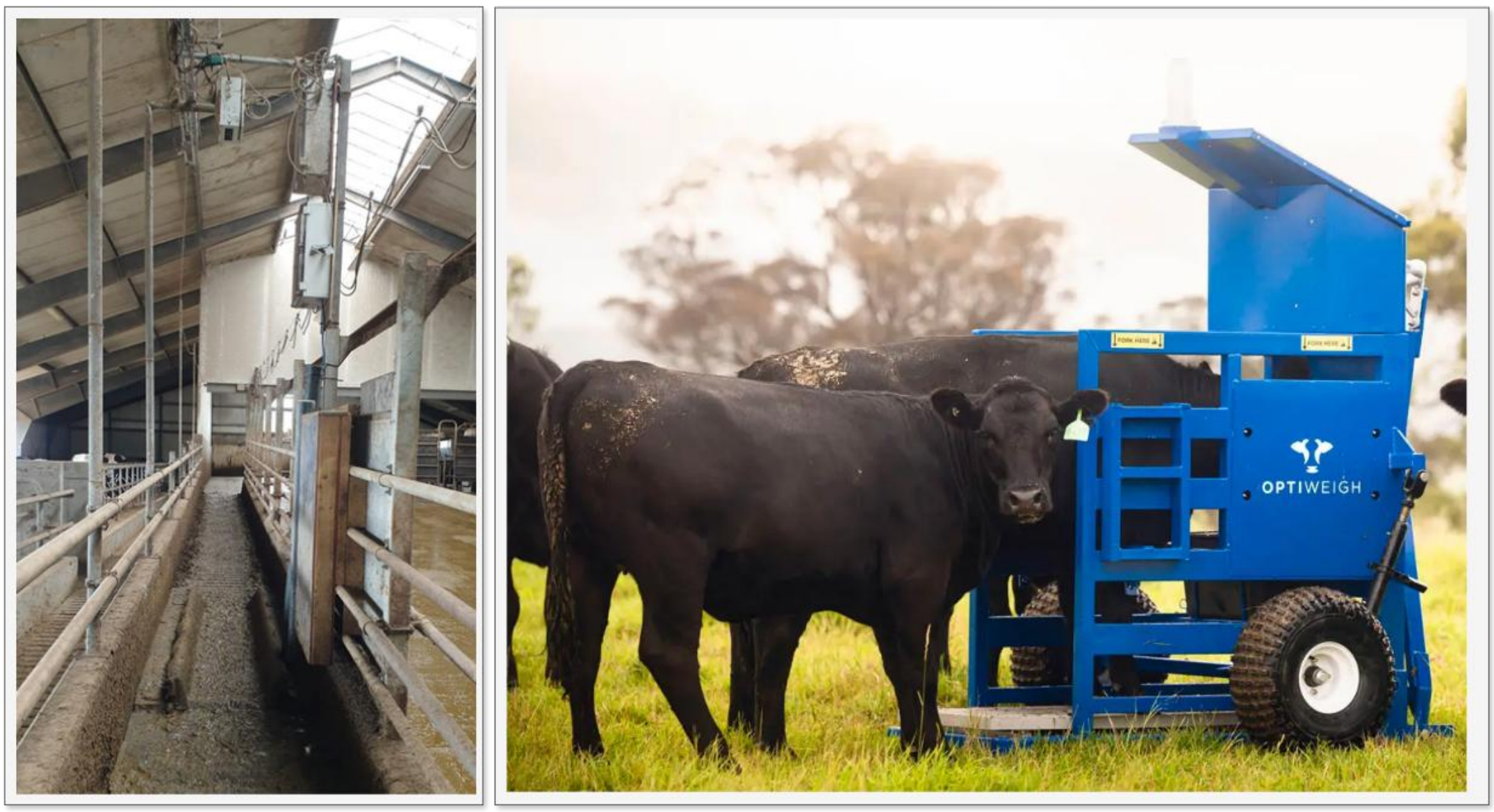

(a) (b)

Fig. 6. (a) Image-based setup: the corridor designed for capturing 3D cattle imagery. (b) Numerical data-based setup: the Optiweight device used for collecting numerical weight data.

Quantitative morphological features involve numeric data, offering precise measurements that capture the dimensions and structural attributes of the cattle. Various scholars have effectively implemented a range of ML models to estimate the body weight gain of cattle using these features (Gjergji et al., 2020; Shahinfar et al., 2020). These studies have demonstrated the capacity of ML algorithms to estimate the non-linear relationship between CWG and quantitative morphological features (Alonso et al., 2013; Bezsonov et al., 2021). Fig. 7 shows the top four most utilised quantitative morphological features, namely Body Length, Heart Girth, Hip Height and Withers Height, along with their respective description. According to the figure, Body Length emerges as the most frequently employed quantitative morphological feature, utilised 13 times for estimating CWG research. The following are Heart Girth, Withers Height, and Hip Height, used 10, 9, and 7 times, respectively. Other quantitative morphological features include Chest Depth (Ruchay, Kober, Dorofeev, Kolpakov, Dzhulamanov, et al., 2022; Vanvanhossou et al., 2018), Chest Girth (Gruber et al., 2018; Grzesiak et al., 2023; Vanvanhossou et al., 2018), Chest Width (Ruchay, Kober, Dorofeev, Kolpakov, Dzhulamanov, et al., 2022), Hip Width (Gruber et al., 2018; Heinrichs et al., 1992), Muzzle Circumference (Wesly et al., 2021), Rump Width (Vanvanhossou et al., 2018), Sacrum Height (Vanvanhossou et al., 2018), and Shank Circumference (Wesly et al., 2021).

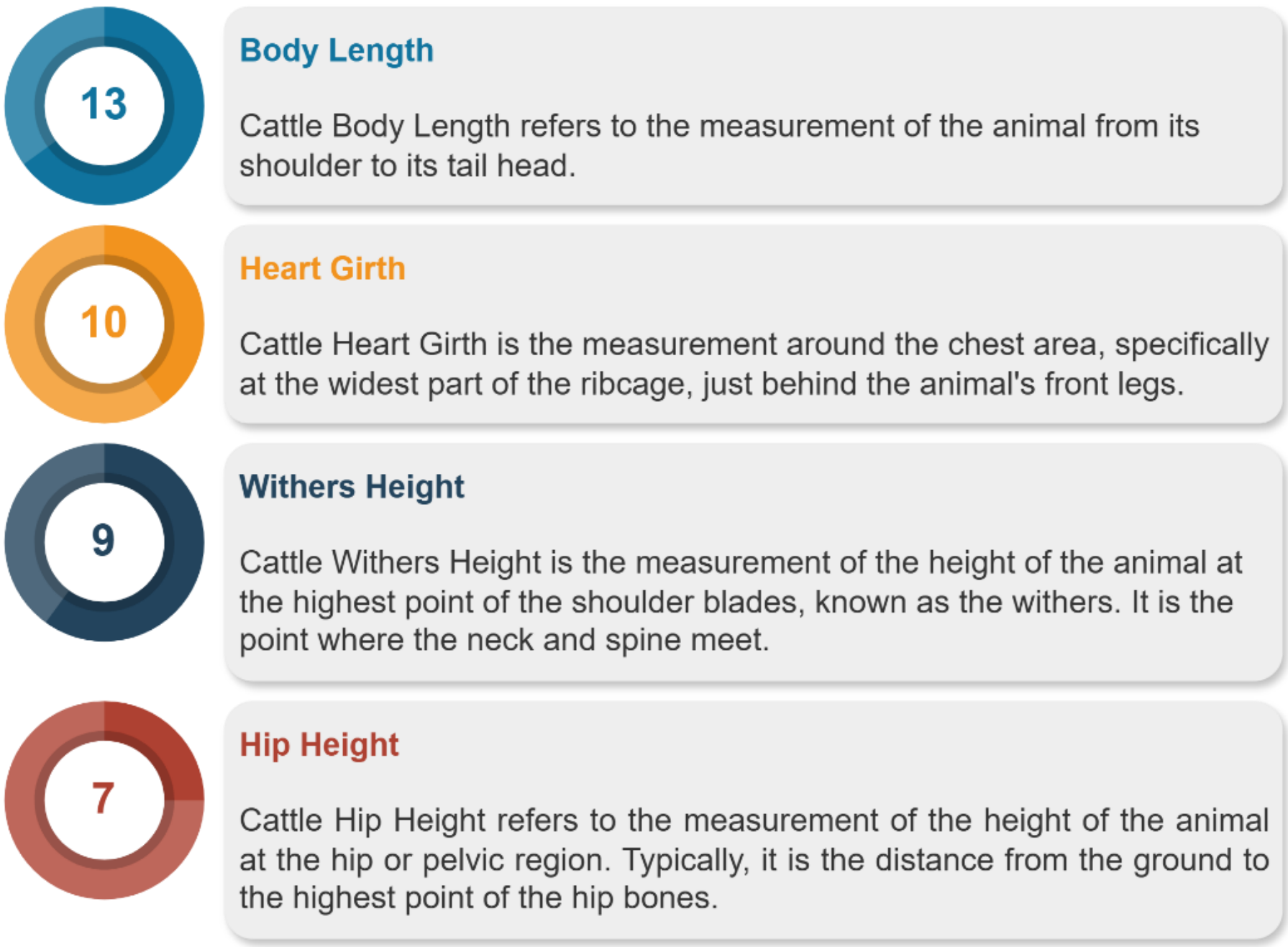

Fig. 7. The top four quantitative morphological features for estimating CWG, with circle numbers indicating usage frequency in papers.

Image-based morphological features present an alternative approach to estimating CWG (Bezen et al., 2020). This method involves the acquisition of cattle images, followed by the utilisation of advanced image processing techniques to evaluate their physical characteristics and overall body condition (Nir et al., 2018). In the realm of estimating CWG research, the collection of image-based measurement features often involves the acquisition of diverse visual data types. In the process of gathering image-based features, researchers commonly utilise four types of images, namely, 2D vision (Afridi et al., 2024; Weber et al., 2020), thermal vision (Stajnko et al., 2008), stereo vision employing calibrated 2D cameras (Tasdemir et al., 2011b), and 3D vision (Gebreyesus et al., 2023; Jang et al., 2020; Lan et al., 2024; Song et al., 2018). Fig. 8 depicts the number of research papers that utilised a set of visual techniques for morphological features to estimate CWG. The figure indicates that 2D vision was employed 15 times, thermal vision once, stereo vision twice, and 3D vision emerged as the most frequently utilised, implemented 21 times in the research papers. These findings highlight the varied preferences for different visual techniques when studying the morphological features linked to estimating CWG.

Table 8 describes the top nine image-based morphological features using computer vision techniques that have garnered significant attention within the research community. Notably, Withers Height emerges as the most frequently employed image-based morphological feature, appearing in nineteen primary research studies. Simultaneously, Body Length and Hip Height emerged as the second most commonly used features, each being included in eighteen studies. The feature Hip Width ranked third in usage frequency and appeared in fourteen individual studies. Additionally, several researchers have explored the utility of other image-based morphological features such as Body Diagonal Variable (de Moraes Weber et al., 2020; Yan et al., 2019), Chest Depth (Bozkurt et al., 2007; Dang et al., 2022; Le Cozler et al., 2019; Ozkaya, 2013), Chest Width (Dang et al., 2022; Jang et al., 2020), Heart Girth (de Moraes Weber et al., 2020; Kaya & Bardakcioglu, 2021; Le Cozler et al., 2019), Length of Nick to Rump(Xiong et al., 2023), and Rump Length (Alonso et al., 2013; Dang et al., 2022; Song et al., 2018), showcasing a diverse array of parameters under investigation in these studies.

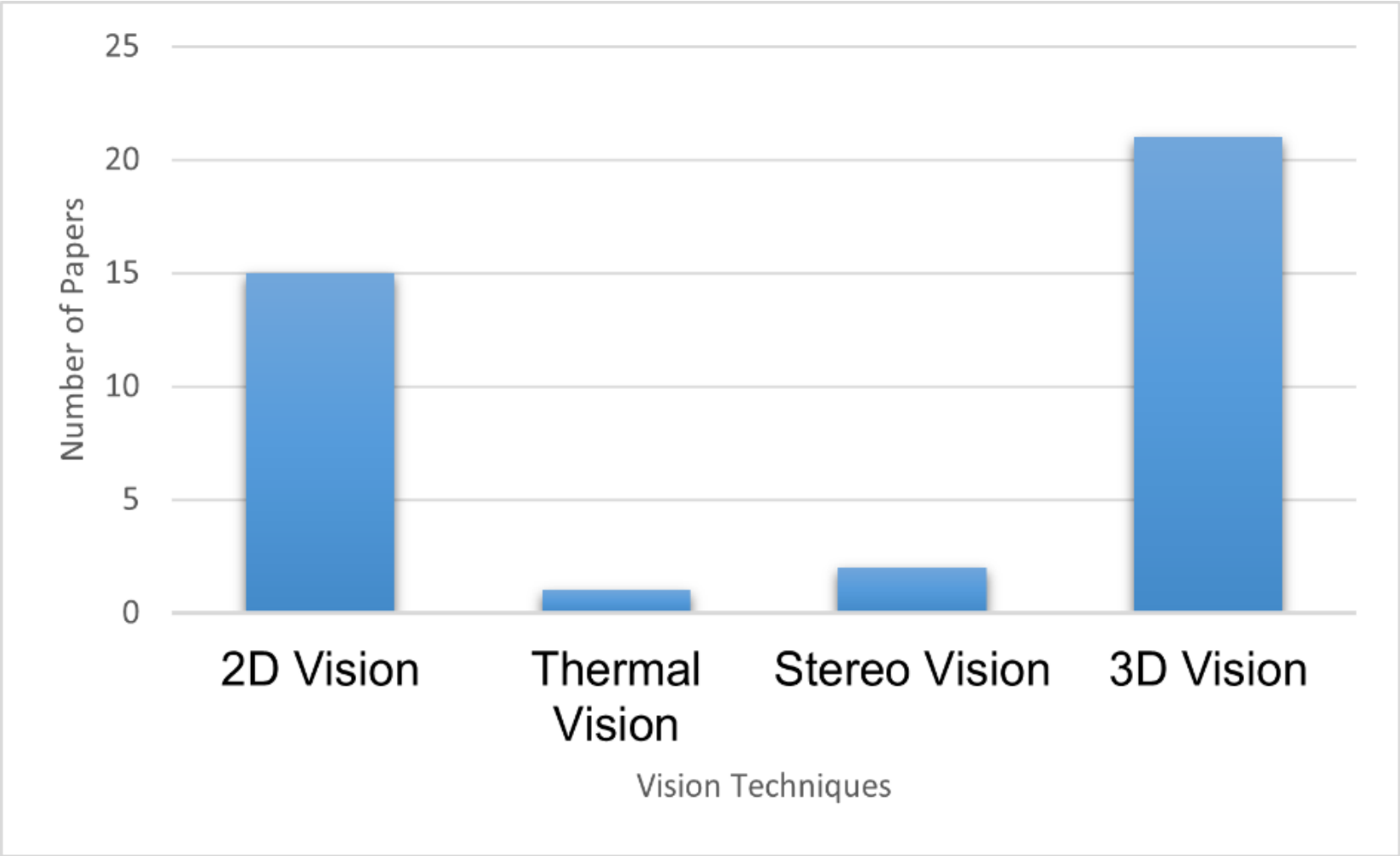

Fig. 8. The vision techniques used in estimating CWG research.

**Table 8**
Top nine image-based features for estimating CWG

| SN | Feature | Description | No. of papers |
|---|---|---|---|
| 01 | Body Width | It is the horizontal measurement across the widest part of a cattle body, typically around the midpoint. | 10 |
| 02 | Body Length | The measurement extends from the front of a cattle's shoulders to the commencement of the tailbone. | 18 |
| 03 | Body Side Area | It is the measurement of the lateral surface of a cattle body. | 08 |
| 04 | Body Volume | It refers to the spatial dimensions encompassed by a cattle's body in three dimensions. | 09 |
| 05 | Chest Girth | The measurement of a cattle's circumference directly behind the front legs provides an assessment of its circumference in that specific region. | 08 |
| 06 | Dorsal Area | The dorsal area encompasses the entire upper part of an organism, often from a cattle's head to the tail. | 09 |
| 07 | Hip Height | The minimum distance from the surface a cattle is standing on to the top of the hips. | 18 |
| 08 | Hip Width | The measurement extends from one hip bone of a cattle to the other. | 14 |
| 09 | Withers Height | The minimum distance from the surface on which a cattle is standing to the ridge between its shoulder blades. | 19 |

In the field of research and analysis, numerous researchers have broadened their studies by integrating supplementary features with image-based or quantitative characteristics. These added elements often encompass a wide range of variables, including Age (Gebreyesus et al., 2023; Grzesiak et al., 2023; Ruchay, Kober, Dorofeev, Kolpakov, Dzhulamanov et al., 2022; Tebug et al., 2018; Vanvanhossou et al., 2018; Xiong et al., 2023), Days in Milk (Kuzuhara et al., 2015; Song et al., 2018), Gender (Sarini et al., 2023; Tebug et al., 2018; Vanvanhossou et al., 2018), Parity (Kuzuhara et al., 2015; Song et al., 2014, 2018), Breed Group (MacNeil et al., 2021; Tebug et al., 2018) and more.

Feature extraction methods refer to techniques and processes used in various fields, especially in ML, computer vision, and data analysis, to identify and select relevant information or characteristics (features) from raw data or input sources. These techniques aim to decrease the number of dimensions in the data while retaining essential information, making it more manageable and suitable for analysis, modelling, or further processing. Feature extraction is necessary for simplifying complex data, highlighting key patterns, and improving the ML model's performance by focusing on the most informative aspects of the data. In the analysis of 53 articles, it is observed that researchers commonly utilised quantitative and image-based feature extraction methods in their studies on estimating CWG.

Image-based feature extraction is integral in estimating CWG, as it encompasses the extraction of relevant information from images or videos of cattle, playing a pivotal role in the process. The extracted information may include details about the sides, objects, colour, light intensity, noise, and others (Pradana et al., 2017). To accomplish this, researchers employed a variety of techniques for feature extraction. However, image datasets often contain noisy images, presenting a challenge to accurate analysis. In addressing this issue, authors in Yamashita et al. (2017) and Song et al. (2014) used a binary image conversion method, representing the cattle body in white and the background in black. This method minimizes noise, removes unnecessary details, and improves accuracy in subsequent analyses and weight gain estimations. Another strategy, demonstrated by the author in Song et al. (2018), involves the implementation of a removal algorithm to eliminate the noise of data in their research. Moreover, researchers in Tasdemir et al. (2011b) have leveraged visual applications to extract morphological features from the captured images.

Quantitative feature extraction needs extracting relevant information from cattle's physical characteristics and measurements to generate quantifiable features for subsequent analysis. Numerous researchers employed quantitative feature extraction in their research on estimating CWG (Sarini et al., 2023). For example, the authors Sarini et al. (2023) and Vanvanhossou et al. (2018) implemented data preprocessing to remove elements with p-values exceeding 0.05. Fig. 9 illustrates the typical image-based and quantitative feature extraction steps.

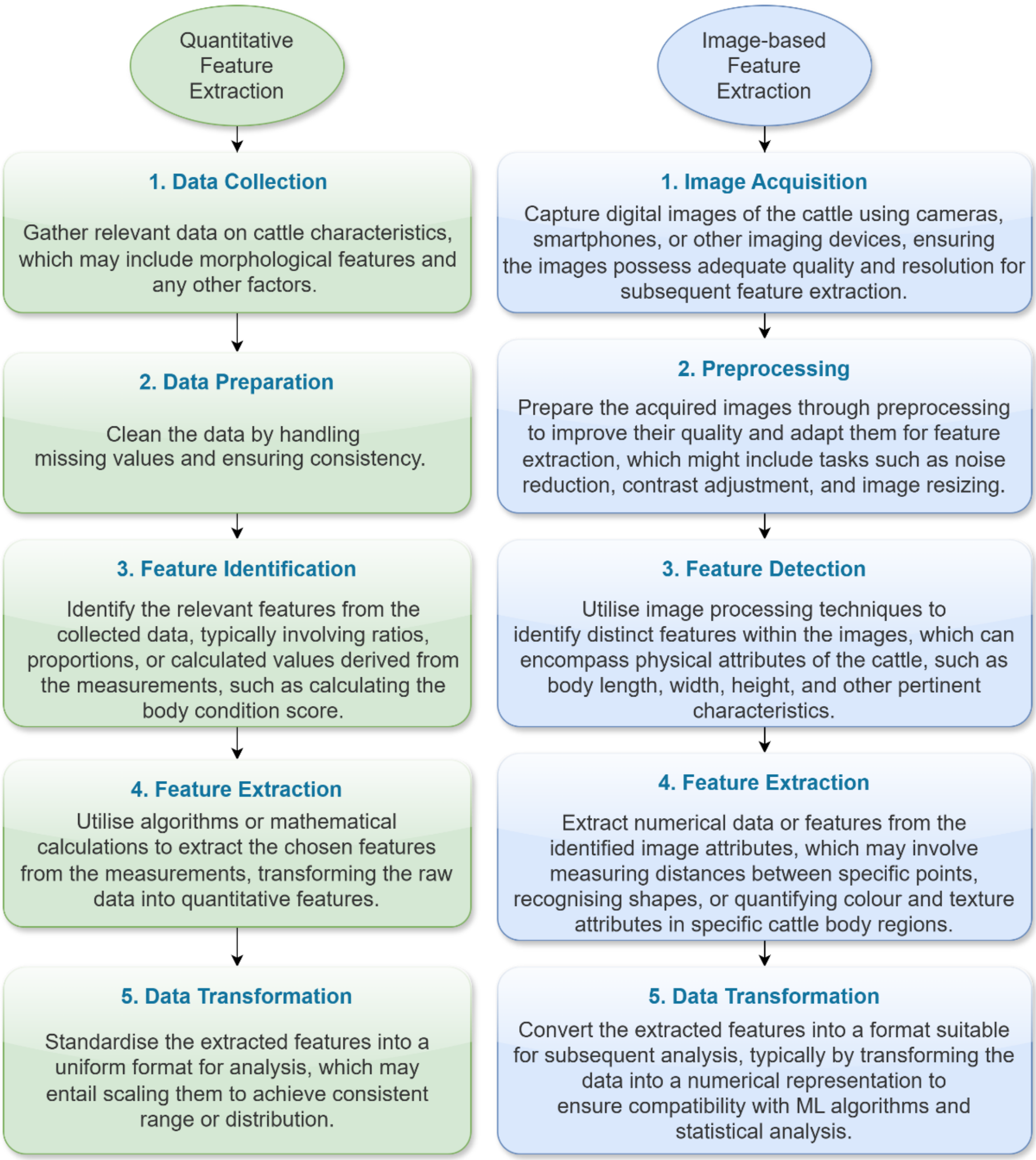

Fig. 9. Typical steps for quantitative and image-based feature extraction.

### *4.3. Evaluation metrics and CV techniques utilised in estimating CWG (RQ3)*

Evaluation metrics contribute significantly to assessing the performance of models by quantifying the disparity between actual and predicted values. In the context of estimating CWG, a specific set of 13 evaluation metrics has been identified to standardize the effectiveness of models. These metrics serve as indispensable instruments for researchers and data scientists, providing a comprehensive understanding of the performance levels attained by their models. The top five evaluation parameters employed in estimating CWG are illustrated in Fig. 10, offering a visual representation of the various criteria used to measure the accuracy and efficiency of the models under scrutiny. Coefficient of Determination ($R^2$) serves as the primary evaluation metric for CWG estimation and is observed 27 times. Root Mean Square Error (RMSE) is the next most frequently employed, appearing 22 times. Mean Absolute Percentage Error (MAPE) is applied 16 times, Correlation Coefficient ($r$) 14 times, and Mean Absolute Error (MAE) rounds out the list with 12 instances. Formula 1 outlines the calculation method for the $r$ evaluation parameter, while Formulas 2 to 4 illustrate the calculation methods for the $R^2$ evaluation parameter. Furthermore, formulas 5, 6, and 7 also showcase the calculation methods for the RMSE, MAE, and MAPE evaluation parameters, respectively. Other evaluation metrics include Average Absolute Error (AAE) (Sarini, N. P., & Dharmawan, 2023), Average Absolute Percentage Error (AAPE) (Sarini, N. P., & Dharmawan, 2023), Intercept (Heinrichs et al., 1992; Sousa et al., 2018; Stajnko et al., 2008), Mean Relative Error (MRE) (Song et al., 2014; Tasdemir et al., 2011a), Mean Square Error (MSE) (Bozkurt et al., 2007; Fuentes et al., 2022; Na et al., 2022; Tasdemir & Ozkan, 2019; Yan et al., 2019), Root Mean Square Error of Prediction (RMSEP) (Cominotte et al., 2020; Gruber et al., 2018; Le Cozler et al., 2019; Liu et al., 2023; Weber et al., 2020), Slope (Fuentes et al., 2022; Sousa et al., 2018; Stajnko et al., 2008), and Sum Square Error (SSE) (Tasdemir et al., 2011a).

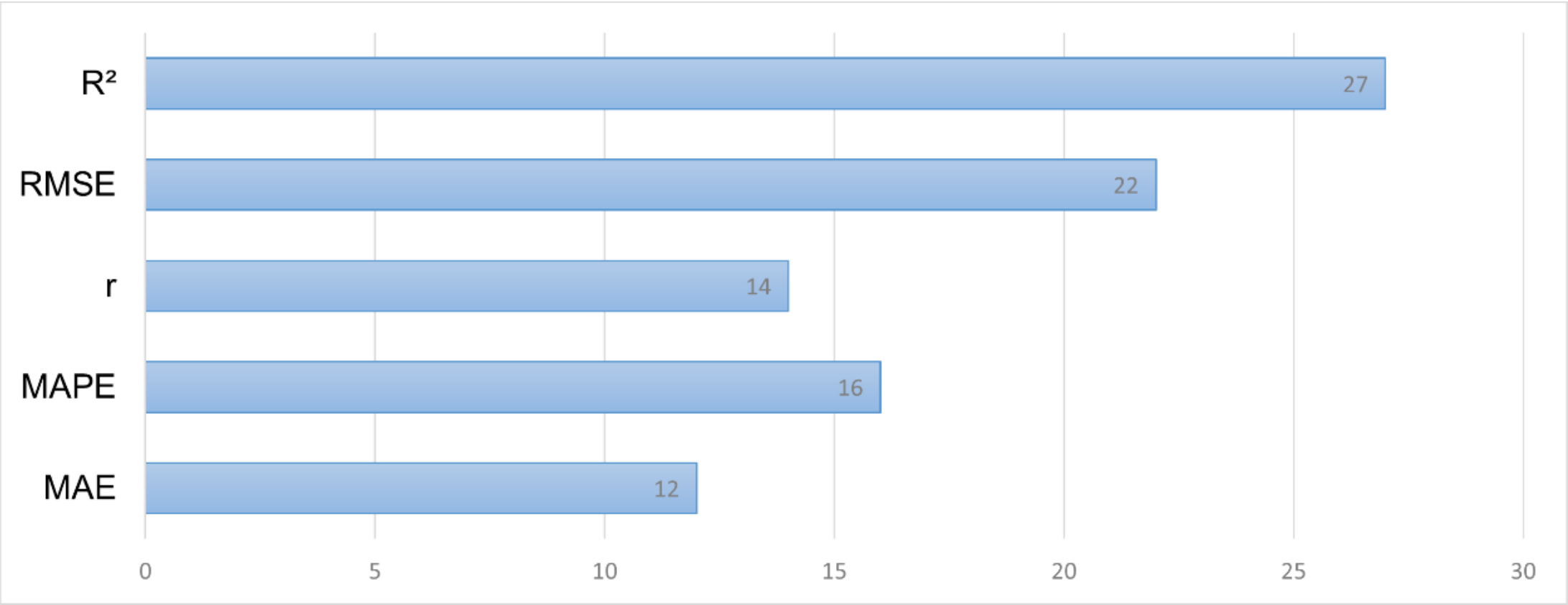

Fig. 10. The distribution of performance metrics employed in selected papers for estimating CWG.

$$r = \frac{n(\sum_{i=1}^{n} y_i \hat{y_i}) - (\sum_{i=1}^{n} y_i)(\sum_{i=1}^{n} \hat{y_i})}{\sqrt{n(\sum_{i=1}^{n} y_i^2) - (\sum_{i=1}^{n} y_i)^2}\sqrt{n(\sum_{i=1}^{n} \hat{y_i}^2) - (\sum_{i=1}^{n} y_i)^2}} \quad (1)$$

Where $y_i$ stands for the actual value and $\hat{y_i}$ represents the estimated value, while $n$ denotes the total record available.

$$R^2 = 1 - \left(\frac{\text{RSS}}{\text{TSS}}\right) \quad (2)$$

$$\text{RSS} = \sum_{i=1}^{n} (y_i - \hat{y_i})^2 \quad (3)$$

$$\text{TSS} = \sum_{i=1}^{n} (y_i - \bar{y})^2 \quad (4)$$

Where TSS represents the sum of squares, RSS indicates the sum of squares for residual, $y_i$ stands for the actual value, $\hat{y_i}$ denotes the estimated value and $\bar{y}$ refers to the average of the actual values.

$$\text{RMSE} = \sqrt{\frac{\sum_{n=1}^{N}(\widehat{r_n} - r_n)^2}{N}} \quad (5)$$

Where $\widehat{r_n}$ signifies the predicted rating, $r_n$ denotes the actual rating in a testing dataset, and $N$ refers to the number of samples in the testing dataset.

$$\text{MAE} = \frac{1}{n}\sum_{i=1}^{n}|y_i - x_i| \quad (6)$$

Where $n$ signified the number of samples in the test dataset, $y_i$ stands for the prediction and $x_i$ represents the actual value.

$$\text{MAPE} = \frac{1}{n}\sum_{i=1}^{n}|(y_i - \widehat{y_i})/y_i| \quad (7)$$

Where $n$ represents the total number of iterations utilised in error calculation, $\widehat{y_i}$ denotes the predicted value and $y_i$ represents the actual value.

CV is a fundamental tool in ML, ensuring reliable model assessment and performance estimation by iteratively using dataset subsets for training and validation. It plays a crucial role in mitigating overfitting, pinpointing optimal hyperparameter values, and ensuring a consistent model performance across diverse data subsets. Despite the existence of various CV methods, this investigation has identified only five meticulous CV methodologies within the specific domain of estimating CWG research. These CV methods encompass the holdout, 2-fold, 5-fold, 10-fold, and leave-one-out approaches. The holdout technique strategically segments the dataset into training, validation, and testing subsets, each with distinct roles in fine-tuning the model and assessing its generalization performance (Kohavi, 1995). On the other hand, when dealing with limited datasets, a k-fold CV proves invaluable as it allows for the division of data into multiple groups determined by the parameter *K*. This method iteratively trains the model on *K-1* subsets while evaluating its performance on the remaining one (Cawley & Talbot, 2008; Kohavi, 1995; Pachouly et al., 2022). The leave-one-out approach takes a different route, using the entire training set at once, aggregating all outcomes to estimate error (Brovelli et al., 2008; Kohavi, 1995). These diverse CV techniques offer a comprehensive toolkit for researchers to analyse and enhance the predictive capabilities of their models in the context of estimating CWG research.

Researchers employed single or multiple CV techniques during their estimation of CWG research. The holdout technique emerged as a predominant choice among researchers conducting estimation of CWG studies. Notably, this approach was frequent in the works of several researchers, such as the authors Afridi et al. (2024), de Moraes Weber et al. (2020), Dang et al. (2022), Fuentes et al. (2022), Gebreyesus et al. (2023), Gjergji et al. (2020), He et al. (2023), Hou et al. (2023), Lan et al. (2024), Lee et al. (2023), Ruchay, Kober, Dorofeev, Kolpakov, Gladkov, et al. (2022), Ruchay, Kober, Dorofeev, Kolpakov, Dzhulamanov, et al. (2022), Sarini et al. (2023), Song et al. (2014), Sousa et al. (2018), Tasdemir & Ozkan, (2019), and Xiong et al. (2023). The 10-fold validation emerged as the second most frequently employed method, as seen in the research conducted by Alonso et al. (2013), Grzesiak et al. (2023), Guvenoglu (2023), Jang et al. (2020), Le Cozler et al. (2019), Miller et al. (2019), Tedde et al. (2021), and Weber et al. (2020). The authors Biase et al. (2022), Cominotte et al. (2020), Song et al. (2018), Sousa et al. (2018), and Cang et al. (2019) preferred the leave-one-out validation. In contrast, the authors Alonso et al. (2013) and Shahinfar et al. (2020) chose the 2-fold validation approach, while the authors Jang et al. (2020) and Xiong et al. (2023) favoured the 5-fold validation. The varying frequencies of these evaluation techniques are illustrated in Fig. 11, providing a clear overview of their usage across the cited studies.

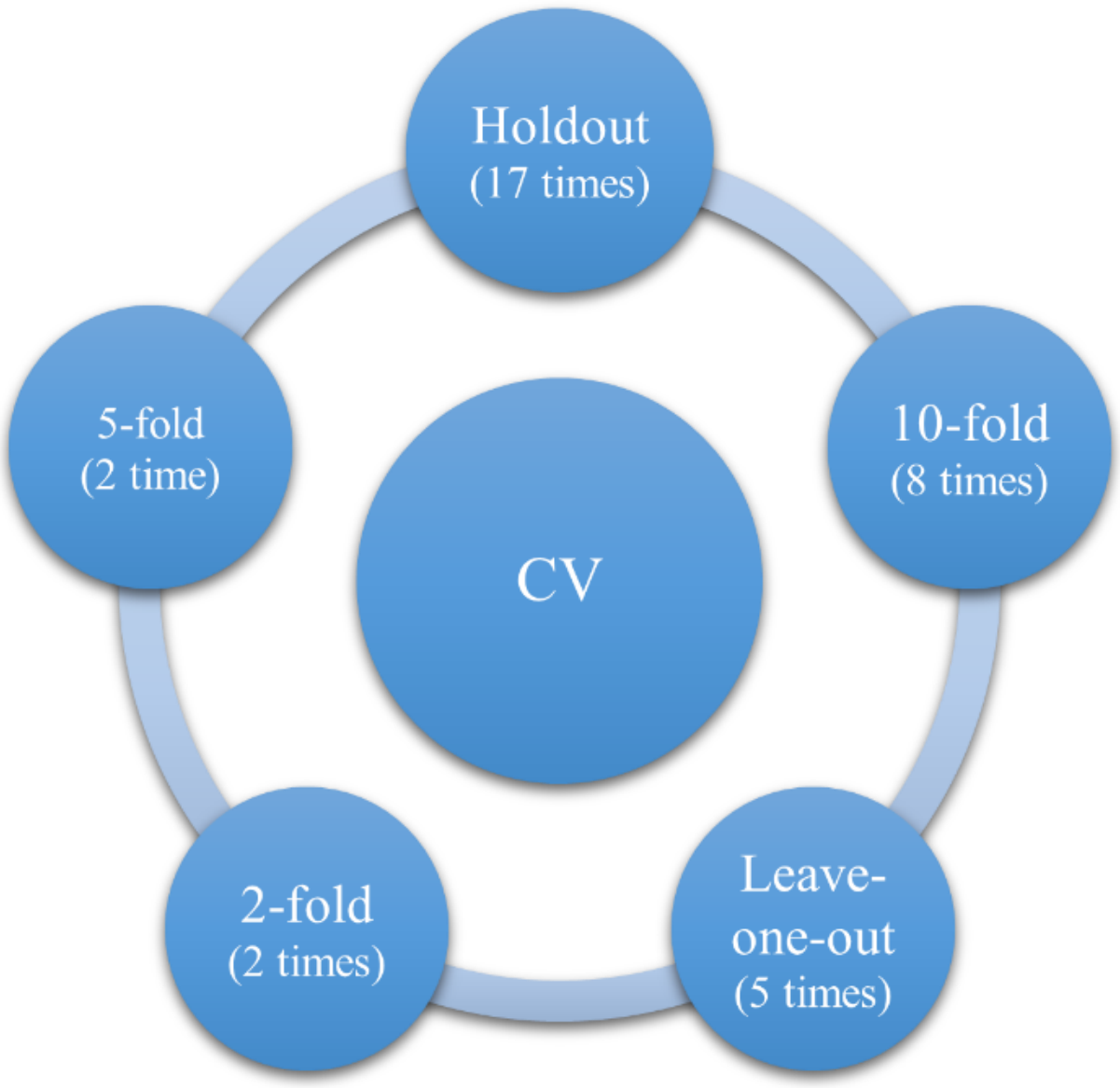

Fig. 11. Five types of CVs for estimating CWG.

### *4.4. Factors of CWG (RQ4)*

CWG is influenced by numerous interconnected factors that collectively impact the profitability of cattle farmers and the availability of high-quality meat products for consumers. There are various equations for estimating CWG, influenced by multiple factors. One commonly utilised formula, Formula 8, illustrates an often-used equation for estimation CWG. Below, several factors have been examined to understand how CWG can be influenced.

$$\mathrm{WG} = \frac{\mathrm{CW-PW}}{\mathrm{dW}} \quad (8)$$

Where WG represents weight gain, cw denotes the current weight, pw indicates the previous weight, and dw marks the days between weightings.

**Genetics:** Genetic factors heavily influence CWG, as these determine each animal's inherent growth capacity, muscle formation, and energy utilisation (Ozkaya, 2013; Weber et al., 2020). Essential genes, including IGF1 (insulin-like growth factor 1), GH (growth hormone), and LEP (leptin), play critical roles in determining growth rate, muscle mass, and fat deposition—traits that significantly impact body composition and growth efficiency (Collares et al., 2014). Cattle, as a diverse group of animals, encompass numerous breeds, each possessing unique genetic predispositions for weight gain (Weber et al., 2020). Due to the genetic diversity among cattle breeds, each breed exhibits unique weight-gain potential. For instance, Angus cattle are well-regarded for their ability to gain weight rapidly.

Researchers have focused on a variety of cattle breeds in the context of estimating CWG. In Fig. 12, comprehensive descriptions are provided for the nine most frequently utilised breeds in research studies focused on estimating CWG. Beyond these focused studies, researchers also explored weight gain research in other cattle types, including Angus (Gjergji et al., 2020; MacNeil et al., 2021; Ruchay, Kober, Dorofeev, Kolpakov, Gladkov, et al., 2022), Ayrshire (Bezsonov et al., 2021; Rudenko et al., 2020), Brown Swiss Cow (Gruber et al., 2018), Fleckvieh (Gruber et al., 2018), Girolando (de Moraes Weber et al., 2020), Hanwoo (Dang et al., 2022; Jang et al., 2020; Na et al., 2022), Hereford Cow (MacNeil et al., 2021; Ruchay, Kober, Dorofeev, Kolpakov, Dzhulamanov, et al., 2022), Krasnaya Stepnaya (Biase et al., 2022), Limousin (Grzesiak et al., 2023), Red Steppe (Rudenko et al., 2020), Somba Cattle (Vanvanhossou et al., 2018), and Zebu Cattle (Tebug et al., 2018). While existing research has broadened our understanding of genetics in CWG, there remains significant scope to investigate further how genetic weight gain ratios vary across breeds.

**Nutrition:** Proper nutrition is a foundation of CWG (Ozkaya, 2013; Weber et al., 2020; Xiong et al., 2023), encompassing essential nutrients such as proteins, carbohydrates, fats, vitamins, and minerals. The composition and quality of the feed directly influence the cattle's ability to gain weight steadily and healthily. Factors such as the energy content of the feed, protein quality, and accessibility to clean water all play a role.

**Environmental Conditions:** Studies have shown that environmental conditions significantly affect CWG (Ozkaya, 2013; Weber et al., 2020; Xiong et al., 2023). These findings demonstrate that cattle of the same breed may display diverse rates of weight gain when exposed to distinct environmental factors (Agung et al., 2018). For example, the authors Agung et al. (2018) pointed out variations in the weight gain of Bali Cattle between Bangka Belitung Province and Bali Island, attributing these differences to the unique environmental conditions in each respective region. Such insights emphasize the crucial role that environmental factors play in shaping CWG and underscore the need for adaptive management strategies that take these conditions into account for effective cattle production and farming.

**Health and Disease Management:** Healthy cattle are more likely to experience consistent and optimal weight gain (Ozkaya, 2013; Weber et al., 2020). Regular veterinary care, vaccination programs, and disease prevention strategies are essential. Parasitic infections, respiratory illnesses, and other health issues can lead to a decreased appetite and weight gain. Timely interventions and appropriate management practices are necessary to address health concerns and minimise their impact on weight gain.

**Cattle Age:** The age of cattle serves as a critical determinant of their weight gain, as explained by Vanvanhossou et al. (2018), who identified age as an influencing factor in the relationships between morphometric measurements and cattle body weight. While young cattle, like calves, typically undergo rapid weight gain due to their ongoing growth, the rate of weight gain may decelerate as they mature (Agung et al., 2018; Gunawan & Jakaria, 2010; Vanvanhossou et al., 2018).

**Cattle Sex:** The sex of cattle, whether male (bulls or steers) or female (heifers or cows), plays a significant role in the context of weight gain, as evidenced by prior studies (Agung et al., 2018; Gunawan & Jakaria, 2010; Vanvanhossou et al., 2018). Extensive documentation supports the observation that bulls in many cattle breeds tend to achieve weight gain faster than heifers, primarily due to differences in hormonal profiles (Kostusiak et al., 2023). A comprehensive understanding of the impact of cattle sex on weight gain holds significant importance for the effective management and optimization of cattle production systems.

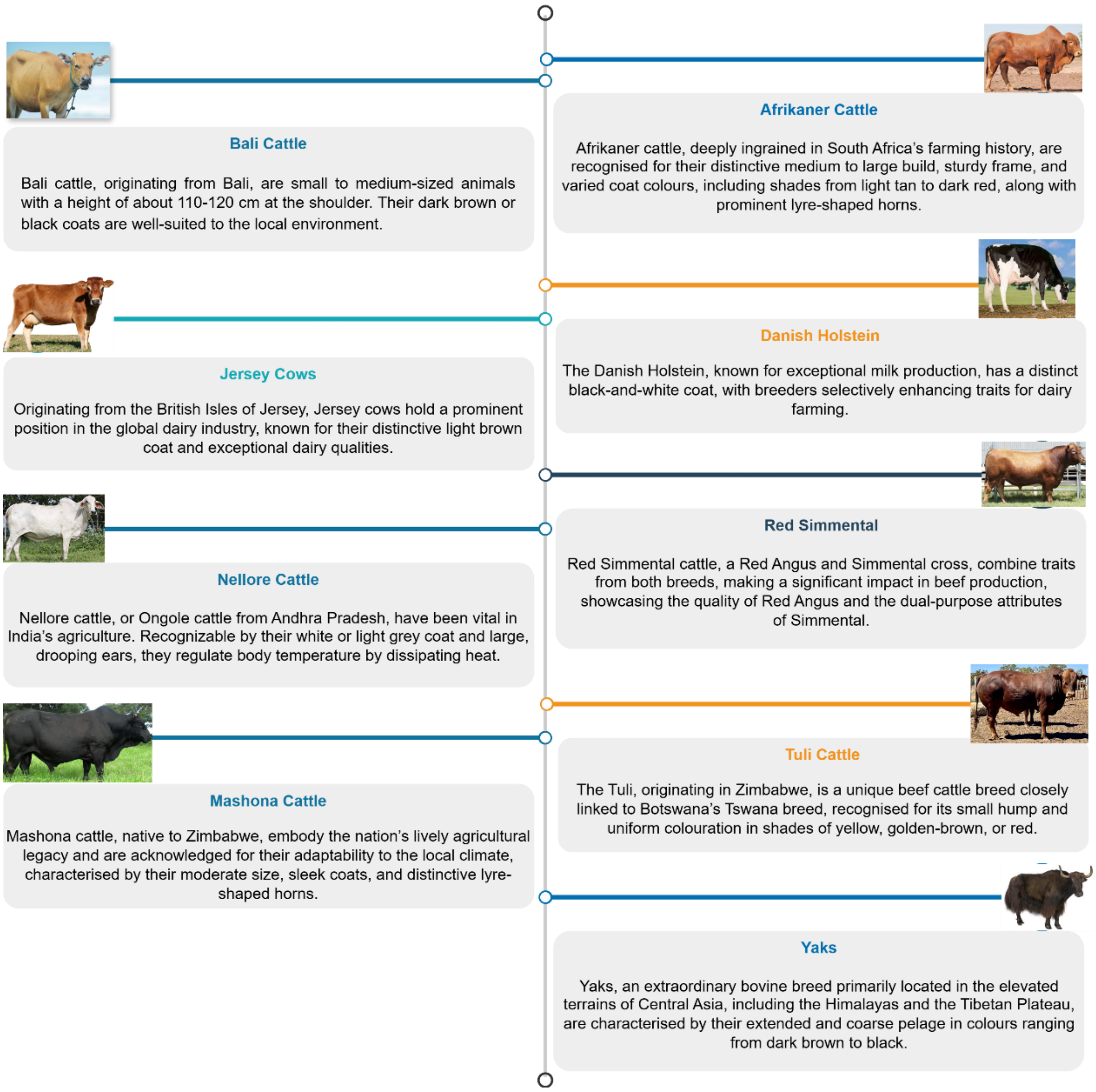

Fig. 12. List of cattle varieties for estimating weight gain.

CWG is intricately influenced by a myriad of factors, including genetics, nutrition, health, environmental conditions, age, and gender, all working together to shape the success of livestock farming. By acknowledging the significance of these elements, farmers can deploy comprehensive strategies that support optimal growth and productivity. Sustainability and efficiency in cattle farming depend on a deep understanding of these elements, which ensures the production of high-quality meat products for consumers and the economic success of the industry.

*4.5. Challenges in the field of estimating CWG (RQ5)*

Estimating CWG using ML and DL techniques presents several challenges due to the complex nature of the domain and the unique characteristics of livestock. The various challenges discussed by researchers in their research articles are detailed below.

**Data Limitation:** One prominent issue highlighted by multiple studies is the insufficient availability of data. Several studies reviewed in this SLR highlighted that small and imbalanced datasets were the leading determinants of reduced accuracy in estimating CWG (Kuzuhara et al., 2015; Shahinfar et al., 2020). Although these studies reported satisfactory performance using limited data, they emphasised the importance of utilising more diverse data for further testing.

**Data Variability:** Estimating CWG data exhibited high variability in the research articles due to variations in breed, age, size, and environmental conditions. For instance, the authors Sarini et al. (2023) noted that differences in the age of the animals accounted for a significant variation in CWG results. The authors Weber et al. (2020) revealed that employing advanced ML techniques on image-based data with identical genetic characteristics of cattle did not yield consistent results when applied to cattle with differing genetic traits. In addition, extracting relevant and consistent features from such diverse data presents a challenging task. Moreover, the limited availability of datasets related to estimating CWG in the public domain further complicates the matter. Although a few databases are publicly accessible (Afridi et al., 2024), most researchers rely on their datasets to estimate CWG. A notable concern is the absence of a benchmark dataset comprehensively covering all relevant features for estimating CWG. The lack of uniform data standards results in ML and DL models relying heavily on specific datasets for estimating CWG, impacting their overall performance. Consequently, achieving generalised and robust estimation models across diverse cattle populations remains a challenge for estimating CWG.

**Data Quality:** Quality data is indispensable for enhancing the accuracy of estimating CWG using advanced ML technology. The analysis of 53 selected articles revealed that most researchers utilised image-based features for estimating CWG. Poor image quality in this context hinders model performance significantly (Le Cozler et al., 2019; Pradana et al., 2017). The presence of noisy images in datasets (Gebreyesus et al., 2023; Yamashita et al., 2017), attributable to factors like weather conditions, lighting, and other environmental noise, is the primary cause of low data quality. Consequently, extracting accurate features becomes difficult in such scenarios. Another factor is the extended processing time required for high-quality image datasets. To tackle this issue, researchers have adopted techniques such as reducing image dimensions or dividing images into smaller pieces to expedite data processing, which inadvertently compromises data quality. It is noteworthy that the dependence on predetermined camera configurations for image-based data introduces a vulnerability (Dohmen et al., 2022). Any post-training modifications to the camera angle or position affect model accuracy, thereby underscoring the need for careful consideration and examination in subsequent research pursuits.

**Data Preprocessing:** Data preprocessing is a vital keystone in the domain of advanced ML technologies, forming the basis for accurate and efficient model training and prediction. Many researchers have grappled with challenges in their pursuit of acceptable data preprocessing methods. For instance, the authors Yamashita et al. (2017) captured motion images of individual cattle in front of the camera for data. However, the labour-intensive nature of moving cattle for imaging parallels the effort needed to use a weighing machine. In response to these challenges, they proposed the development of a method for extracting analysed images from motion videos of cattle residing in barns. The authors Weber et al. (2020) also recommended utilising video frames capturing the positions of cattle that more effectively elucidate weight-related measurements. This innovative approach aims to boost the accuracy of weight gain estimation more practically and efficiently within the realm of advanced ML technologies.

**Time-series Data:** Time-series data for estimating CWG refers to a sequential and chronological collection of observations or measurements related to the growth and development of cattle over time. This type of data serves as a crucial resource for researchers immersed in estimating CWG. In estimating CWG research by (Gebreyesus et al., 2023), time-series data unveiled significant challenges, illustrating substantial variability in the body weight measurements of cattle, with some individuals having extensive records across different periods and others possessing only one or a few records. This variation was particularly conspicuous in the body weight measurements of cows at early ages compared to later stages of development. These findings highlight the intricate and diverse nature of time-series data, underscoring the necessity for nuanced analytical approaches to yield accurate and comprehensive insights into the complexities of estimating CWG.

**Feature Selection:** The process of feature selection in the context of estimating CWG entails several challenges attributable to the intricate nature of the underlying biological and environmental determinants influencing weight gain in cattle. A primary challenge in this endeavour is the judicious selection of pertinent features, a task of critical importance to circumvent overfitting and augment the interpretability of the predictive model (Alonso et al., 2013). The appropriateness of selected features is contingent upon the inherent heterogeneity within cattle populations, encompassing diverse breeds and variable environmental conditions. The presence of extraneous and redundant features introduces further intricacies into the feature selection process, necessitating meticulous consideration of feature relevance and interdependencies. Furthermore, the dynamic nature of factors such as seasonal variations, alterations in diet, and fluctuations in health conditions introduces additional complexity, making it difficult to identify stable and consistent features for predictive modelling.

Addressing these challenges requires careful consideration of data representation, model architecture, regularization techniques, and continuous monitoring and validation of model performance on different cattle populations and farm environments. Advanced ML techniques, such as ensemble methods and data augmentation, may also aid in improving model generalisation for estimating CWG.

## 5. Our findings and future research directions

The review paper identifies several findings and future research needs. The various findings and future research directions are detailed below.

**Dominance of LR with Shifting Preference Toward Advanced Models and Increased Adoption of DL and Image-Based Approaches:** LR continues to dominate CWG estimation models due to its simplicity and effectiveness, reflected in its use in 33 out of 53 reviewed studies. However, there is a noticeable shift toward more advanced models like ANN, SVM, and RF, which excel at capturing complex, non-linear relationships in the data. Additionally, DL techniques, particularly CNNs and MLPs, are gaining traction for CWG estimation, especially in image-based contexts. These DL models deliver higher accuracy and adaptability compared to traditional methods, underscoring a clear trend toward leveraging sophisticated, data-driven approaches for more precise CWG estimation.

**Diversity of Feature Types, Importance of Data Preprocessing, and Integration of Supplementary Features:** Research on CWG estimation frequently incorporates quantitative features, such as body length and heart girth, along with image-based features, allowing for a more robust model by combining precise numerical data with advanced image processing techniques. The quality of these features relies heavily on data preprocessing steps like noise reduction, outlier removal, and binary image conversion, all of which are essential for effective feature extraction and model accuracy. Additionally, many studies integrate supplementary variables, such as age, gender, and days in milk, alongside morphological data, significantly enhancing predictive power by providing a comprehensive view of the factors influencing cattle weight gain.

**Dominant Evaluation Metrics and Variety of Cross-Validation Techniques:** Evaluation metrics and CV techniques play a pivotal role in CWG model assessment, with $R^2$ emerging as the most frequently applied metric (27 instances) due to its importance in measuring explained variance. RMSE is the second most common metric (22 instances) for evaluating prediction error. In contrast, MAPE (16 instances), $r$ (14 instances) and MAE (10 instances) are also widely used, indicating a strong emphasis on both accuracy and error transparency. For CV, the holdout method leads in popularity due to its simplicity. However, the 10-fold CV method follows closely, providing a balanced approach to bias and variance, which is beneficial for studies with limited data. Together, these metrics and CV techniques ensure thorough validation and reliability in CWG estimation models.

1. **Six impactful factors on CWG**: Genetics plays a key role, with breeds like Angus, Ayrshire, Limousin, and Zebu showing different weight gain potentials. Proper nutrition, including balanced proteins, carbohydrates, fats, and minerals, is vital for consistent growth, while environmental conditions also affect growth rates. Effective health management, involving disease prevention and veterinary care, further supports optimal weight gain. Younger cattle tend to grow faster, and males generally gain weight more quickly than females due to hormonal differences. Therefore, successful cattle farming requires a holistic approach to managing these interconnected factors—genetics, nutrition, environment, health, age, and sex—to ensure sustainability and economic viability.

1. **Six challenges in estimating CWG research**: This survey highlights critical challenges in estimating CWG, including data limitations, variability, quality, preprocessing, and feature selection. A significant issue is the lack of diverse datasets, which reduces accuracy and leads to inconsistent results across breeds. High variability is caused by breed, age, and environmental conditions, while poor image quality and long processing times hinder model performance. Time-series data variability requires advanced analytical methods, and complex biological factors complicate feature selection, increasing the risk of overfitting. Advanced machine learning techniques, such as ensemble methods and data augmentation, are vital for improving CWG estimation.

**Future research directions**: A total of 53 research papers have been reviewed on the estimation of CWG, yet important gaps still exist. Specifically, there is limited research dedicated to predictive or forecasting approaches for future weight gain. Predictive modelling, which differs from estimation, could offer farmers substantial advantages by forecasting growth patterns, optimizing resources, and enhancing profitability. Most current research emphasizes individual CWG, but there is no significance research on mob-based or group prediction approaches to better align with the global rise in mob-based cattle farming. This transition may yield more scalable and resilient solutions. While factors like sex and age have been studied, more research is needed to develop flexible models that adapt to a broader range of factors, including diet, environmental conditions, and health, for accurate tracking of CWG over time. The impact of climate change on CWG is also crucial, as changes in temperature, humidity, and pasture quality could affect the growth of cattle. It is also suggested that diverse data about cattle, which includes various breeds, ages, and other factors, be integrated to extend the effectiveness of predictive models.

## 6. Conclusions

This paper underscores the advanced ML techniques for estimating CWG, showcasing a comprehensive assessment of model performance and feature relevance. Performance comparisons are provided in Table 5, while Table 6 summarizes each model's strengths and limitations, creating a practical guide for researchers. Section 4.2 discusses the significance of targeted feature extraction and selection in improving model precision and minimizing data loss. Section 4.3 addresses the various evaluation metrics and CV techniques applied in CWG estimation, showcasing the range of assessment methods. Section 4.4 examines influential factors in CWG, providing deeper insights into the elements affecting estimation accuracy. Finally, Section 4.5 identifies significant challenges in the field, encouraging further research to address current gaps and enhance estimation capabilities in CWG. In light of these findings, this study recommends predictive modelling for forecasting future weight gain, which is essential for efficient resource utilization in cattle farming. The study further recommends exploring mob-based cattle weight prediction using advanced ML methods, offering significant advantages for large-scale farming enterprises. These approaches open new research avenues to estimate and forecast CWG, supporting the livestock industry while contributing to sustainable agricultural practices.

## Funding sources

This project was supported by funding from Food Agility CRC Ltd under the Commonwealth Government CRC Program. The CRC Program supports industry-led collaborations between industry, researchers, and the community.

## Declaration of competing interest

The authors declare that they have no known competing financial interests or personal relationships that could have influenced the work reported in this paper.